\documentclass[sigconf]{acmart}
\AtBeginDocument{%
  \providecommand\BibTeX{{%
    \normalfont B\kern-0.5em{\scshape i\kern-0.25em b}\kern-0.8em\TeX}}}

\copyrightyear{2022}
\acmYear{2022}
\setcopyright{rightsretained}
\acmConference[FAccT '22]{2022 ACM Conference on Fairness, Accountability, and Transparency}{June 21--24, 2022}{Seoul, Republic of Korea}
\acmBooktitle{2022 ACM Conference on Fairness, Accountability, and Transparency (FAccT '22), June 21--24, 2022, Seoul, Republic of Korea}\acmDOI{10.1145/3531146.3533168}
\acmISBN{978-1-4503-9352-2/22/06}

\usepackage[utf8]{inputenc}

\usepackage{graphicx}
\graphicspath{ {images/} }

\usepackage{amsthm} 
\usepackage{url} 
\usepackage{bm} 
\usepackage{enumitem} 
\usepackage{hyperref} 
\usepackage{bbm} 

\usepackage{soul}

\usepackage{multirow}

\usepackage{xspace} 

\usepackage{algorithm} 
\usepackage{algpseudocode} 

\usepackage[colorinlistoftodos,prependcaption,textsize=small]{todonotes} 

\usepackage{widetext}

\usepackage{lipsum}
\usepackage{mathtools}
\usepackage{cuted}

\definecolor{cadmiumgreen}{rgb}{0.0, 0.45, 0.26}
\definecolor{lightcadmiumgreen}{rgb}{0.0, 0.60, 0.30}
\definecolor{cadmiumorange}{rgb}{0.93, 0.53, 0.18}
\definecolor{burgundy}{rgb}{0.50, 0.0, 0.13}
\definecolor{airforceblue}{rgb}{0.36, 0.54, 0.66}

\iffalse
    \newcommand{\del}[1]{\colorlet{saved}{.}\color{olive}{#1}\color{saved}\xspace}
    \newcommand{\itodo}[1]{\colorlet{saved}{.}\color{magenta}\textbf{TODO}: {#1} \color{saved}\xspace}
    \newcommand{\todox}[1]{\todo[inline,linecolor=magenta,backgroundcolor=magenta!25,bordercolor=magenta]{\color{magenta}\textbf{TODO}: \color{black} #1}}
    
    \newcommand{\EA}[1]{\colorlet{saved}{.}\color{black}{#1}\color{saved}\xspace}
    \newcommand{\EAB}[1]{\colorlet{saved}{.}\color{black}{#1}\color{saved}\xspace}

    \newcommand{\delEAC}[1]{}
    \newcommand{\delEAD}[1]{\colorlet{saved}{.}\color{burgundy}\st{#1}\color{saved}\xspace}

    \newcommand{\EB}[1]{\colorlet{saved}{.}\color{black}{#1}\color{saved}\xspace}
    \newcommand{\EC}[1]{\colorlet{saved}{.}\color{black}{#1}\color{saved}\xspace}
    \newcommand{\ED}[1]{\colorlet{saved}{.}\color{black}{#1}\color{saved}\xspace}
    \newcommand{\EE}[1]{\colorlet{saved}{.}\color{black}{#1}\color{saved}\xspace}
    \newcommand{\EF}[1]{\colorlet{saved}{.}\color{cadmiumgreen}{#1}\color{saved}\xspace}
\else
    \newcommand{\del}[1]{}
    \newcommand{\itodo}[1]{}
    \newcommand{\todox}[1]{}

    \newcommand{\EA}[1]{{#1}\xspace}
    \newcommand{\EAB}[1]{{#1}\xspace}

    \newcommand{\delEAC}[1]{}
    \newcommand{\delEAD}[1]{}

    \newcommand{\EB}[1]{{#1}\xspace}
    \newcommand{\EC}[1]{{#1}\xspace}
    \newcommand{\ED}[1]{{#1}\xspace}
    \newcommand{\EE}[1]{{#1}\xspace}
    \newcommand{\EF}[1]{{#1}\xspace}
\fi

\makeatletter
\newcommand{\customlabelnref}[2]{\protected@write \@auxout {}{\string \newlabel {#1}{{#2}{\thepage}{#2}{#1}{}}}\hypertarget{#1}{#2}}
\newcommand{\customlabel}[2]{\protected@write \@auxout {}{\string \newlabel {#1}{{#2}{\thepage}{#2}{#1}{}}}}
\makeatother

\newcommand{\norm}[1]{\left\lVert#1\right\rVert} 

\DeclareMathOperator*{\argmax}{arg\,max}
\DeclareMathOperator*{\arginf}{arg\,inf}
\newcommand\Tau{\mathrm{T}}
\newcommand{\m}{m}
\newcommand{\F}{\mathcal{F}}
\newcommand{\X}{\mathcal{X}}
\newcommand{\Y}{\mathcal{Y}}

\newcommand{\x}{\bm{x}}
\newcommand{\af}{{A}}
\newcommand{\cf}{{c}}
\newcommand{\df}{{CF}}
\newcommand{\pf}{{p}}

\newcommand{\apropk}{\tilde{A}_{k}}
\newcommand{\arandk}{\bar{A}_{k}}
\newcommand{\xp}{\bm{x'}}
\newcommand{\R}{\mathbb{R}}
\newcommand{\NNZ}{\mathbb{N}_{>0}}

\newcommand{\tauv}{\bm{\tau}}
\newcommand{\phiv}{\bm{\phi}}

\newcommand{\Ifunc}{\vmathbb{I}^k}

\newcommand{\D}{\mathcal{D}}
\newcommand{\N}{\mathcal{Q}}
\newcommand{\Phiv}{{\Phi}}
\newcommand{\Tauv}{{\Tau}}

\newcommand{\E}{\mathbb{E}}
\newcommand{\RD}{\mathcal{R}}
\newcommand{\rv}{\bm{r}_{\RD}}
\newcommand{\rvar}{R}
\newcommand{\Ivec}{\Ifunc[\phiv]}
\newcommand{\Iveci}{\Ifunc_i[\phiv]}
\newcommand{\BigOne}{\mathbbm{1}}
\newcommand{\Indicator}[1]{\BigOne\!\left[ #1 \right]}
\newcommand{\NN}{N\!\!\!N}

\makeatletter
\newcommand{\spaceornewline}[1]{%
\if@twocolumn%
$$
$$
\else
#1
\fi
}%
\makeatother

\begin{document}

\title{Counterfactual Shapley Additive Explanations}

\author{Emanuele Albini}
\orcid{0000-0003-2964-4638}
\affiliation{%
  \institution{J.P. Morgan AI Research}
  \city{London}
  \country{UK}
}
\email{emanuele.albini@jpmorgan.com}

\author{Jason Long}
\affiliation{%
  \institution{J.P. Morgan AI Research}
  \city{London}
  \country{UK}
}
\email{jason.x.long@jpmorgan.com}

\author{Danial Dervovic}
\affiliation{%
  \institution{J.P. Morgan AI Research}
  \city{London}
  \country{UK}
}
\email{danial.dervovic@jpmorgan.com}

\author{Daniele Magazzeni}
\affiliation{%
  \institution{J.P. Morgan AI Research}
  \city{London}
  \country{UK}
}
\email{daniele.magazzeni@jpmorgan.com}

\renewcommand{\shortauthors}{Albini, et al.}

\begin{abstract}

Feature attributions are a common paradigm for model explanations due to their simplicity in assigning a single numeric score for each input feature to a model.
In the actionable recourse setting, wherein the goal of the explanations is to improve outcomes for model consumers, it is often unclear how feature attributions should be correctly used.
With this work, we aim to strengthen and clarify the link between actionable recourse and feature attributions.
Concretely, we propose a variant of SHAP, \emph{Counterfactual SHAP} (CF-SHAP), that incorporates counterfactual information to produce a \emph{background dataset} for use within the marginal (a.k.a. interventional) Shapley value framework.
We motivate the need within the actionable recourse setting for careful consideration of background datasets when using Shapley values for feature attributions with numerous synthetic examples.
Moreover, we demonstrate the efficacy of CF-SHAP by proposing and justifying a quantitative score for feature attributions, \emph{counterfactual-ability}, showing that as measured by this metric,
CF-SHAP is superior to existing methods when evaluated on public datasets using tree ensembles.


\end{abstract}




\maketitle

\section{Introduction}\label{sec:intro}


Government regulators are placing increasing emphasis on the fairness and discrimination issues in decision making processes using machine learning algorithms in high-stakes context as finance and healthcare.  For example, \EF{the European Commission \cite{EuropeanCommission2019} put particular emphasis on the right to explain AI systems decisions while the U.S. credit regulations \cite{EqualOpportunityAct} are even more specific as they prescribe that automatic decisions must be explained in terms of key factors that contributed to an adverse decision}.
\EF{At the same time}, in the academic literature several techniques have been proposed to address this issue (see \cite{BarredoArrieta2020,Guidotti2019a,Adadi2018,Cyras2021} for an overview).

In the context of \emph{local explainability} many approaches on which researchers have focused in the last years are based on the notion of \emph{feature attribution}, i.e., distributing the output of the model for a specific input to its features (e.g., \cite{Ribeiro2016,Lundberg17,Merrill2019}).
In this paper in particular we will focus on SHAP, one of the most popular techniques to generate local explanations based on the notion of Shapley value \cite{Shapley1951} from game theory.
Shapley value-based frameworks for Explainable AI (XAI) consider each feature as a player in a $m$-person game to fairly distribute the contribution of each feature to the output of the model. To do so they compare the output of the (same) model when a feature is present with that of when the same feature is missing. There are two main limitations with this approach that have been raised in the literature:
\begin{enumerate}[label=(\alph*)]
	\item It is not clear how to define the output of the model when a feature is missing. The most common approach is to estimate it as an expectation over a background distribution of the input features \cite{Merrick2020}.
    \item There is no explicit guidance provided on how a user might alter one’s behavior in a desirable way \cite{Kumar2020}.
\end{enumerate} 

Another popular area of research has developed around \emph{counterfactual explanations}, also known as \emph{algorithmic recourse}, i.e., given a specific input one must find the ``closest possible world (input)'' \cite{Wachter2017} that gives rise to a different outcome. 
In practice, this means that these approaches aim to find (one or more) points that are (1) close to the one we want to explain; and (2) ``plausible'' (where plausibility can be defined in different ways in the literature, see \cite{Keane2021} for more insights). Counterfactual explanations have two main limitations:
\begin{enumerate}[label=(\alph*)]
	\item Most of the approaches in the literature are limited at finding a single counterfactual point. While this may give the user a clear understanding of what they could do in order to reverse an adverse outcome, it does not allow them to choose changes that are more suited for them.
    \item While there has been some attempt at generating diverse sets of counterfactuals (e.g., \cite{Mothilal2020a,Russell2019}), there is no consensus on how to limit the cognitive load for the user caused by the sheer amount of information that is provided, or -- in other words -- on how to provide a more amenable explanation (in terms of size), as advocated also from a social science perspective \cite{Miller2019}.
\end{enumerate}

In this paper we present how these two general approaches for explainability can be combined in order to provide a \emph{counterfactual feature attribution} grounded on the game-theoretic approach afforded by Shapley values that we call \emph{Counterfactual SHAP (CF-SHAP)}.
We are motivated by the desire to retain the simple form of explanation provided by feature attributions, while introducing the actionability properties of counterfactual explanations.



In particular, our contributions are as follows.
\begin{itemize}\setlength\itemsep{0em}
    \item We enumerate the assumptions that are necessary to interpret Shapley values in a counterfactual sense and discuss what it means for a feature attribution method to demonstrate counterfactual behaviour.
    \item We introduce a \EB{general framework to measure the \emph{counterfactual-ability}} of a feature attribution as a way to quantitatively evaluate its ability to suggest to the user how to act upon the input in order to overcome an adverse prediction.
    \item \EE{In order to achieve higher counterfactual-ability, we} propose to \EE{(1)} use (a uniform distribution over) a set of counterfactuals as the background distribution for the computation of Shapley values \EE{and (2) to enrich the explanation with guidance on the direction in which to change the features}, yielding the CF-SHAP algorithm.
    \item \EE{We benchmark CF-SHAP against baseline feature attribution techniques. CF-SHAP, using $100$-nearest neighbours as a simple counterfactual generation technique, is} shown to have the best counterfactual-ability on \EE{3} \EE{publicly available datasets}.
\end{itemize}

We note that in this paper we concentrate on tree-based models for the following reasons: (1) in the context of classification and regression for tabular data, tree-based ensemble models as XGBoost, CatBoost, LightGBM and Random Forest are deemed as the state-of-the-art in terms of performance \cite{Shwartz-Ziv2021} and therefore are widely adopted in many industries including finance \cite{Sudjianto2021Podcast}; (2) interventional Shapley values can be computed exactly \EB{and efficiently} for tree-based models using the algorithm proposed in \cite{Lundberg2020Trees}.



\section{Background}\label{sec:background}

In the remainder of this paper we consider a \EB{trained} binary classification \emph{model} $f : \X \rightarrow \Y$ where $\X = \R^m$ and $\Y = \R$. We define the \emph{decision function} $F : \X \rightarrow \{0, 1\}$ \EB{with \EC{(binary)} \emph{decision threshold} $t \in \R$} \EB{as\footnote{We use lower-case bold symbols to indicate vectors \EB{(e.g., $\x$) and non-bold symbols to indicate scalars (e.g., $x_i$)}.}
$$F(\x) = \begin{cases} 1 & \text{if } f(\x) > t\\ 0 & \text{otherwise}\end{cases}.$$}

We refer to $f(\x)$ as the model \emph{output} and to $F(\x)$ as the model \emph{prediction} or \emph{outcome}. 

\EB{Note that, without loss of generality, if not otherwise specified we use $t = 0$ as decision threshold. 
}
Moreover, without loss of generality, we assume that an input $\x \in \X$ such that $F(\x) = 1$ is an \emph{adverse outcome} for the user, \EB{e.g., the rejection of a loan application}.
We also note that the results in this paper can be trivially generalized to multi-class models.


\subsection{Shapley values}\label{sec:background-shapley}


The Shapley values method is a technique used in classic game theory to fairly attribute the payoff to the players in an $\m$-player cooperative game.
\EE{Given a set of players $\F = \{ 1, \ldots, \m \}$ and the \emph{characteristic function} $v : 2^{\F} \rightarrow \R$ of a game $\Gamma$ Shapley values fairly attribute the payoff returned by the characteristics function to each player.}

In the context of machine learning models the players are the features of the model and several ways have been proposed to simulate feature absence in the characteristic function (e.g., retraining the model without such feature \cite{Strumbelj2010}).
In this paper we use the approximation of the characteristic function proposed in \cite{Lundberg17} and \cite{Lundberg2020Trees} (SHAP) that simulates the absence of a feature using the marginal expectation over a background distribution $\D$.

\EE{Formally, 
the Shapley value of player $i$ is defined as:}
$$
\phi_i = \ED{\frac{1}{\m}} \sum_{S \subseteq \F \setminus \{ i \}} \binom{\m-1}{|S|}^{-1}\left[ v(S \cup \{i \}) - v(S) \right] \spaceornewline{\quad} \text{ where } v(S) = \E_{\xp \sim \D}\left[ f(\x_{S}, \xp_{\F \setminus S}) \right]
$$

\EB{where $\E_{\xp \sim \D}$ denotes the expected value under the distribution $\D$} and with an abuse of notation $f(\x_{S}, \xp_{\F \setminus S})$ indicates the output of the model with feature values $\x$ for features in $S$ and values $\xp$ for feature values not in $S$.


We will henceforth refer to the space of Shapley values $\R^{\m}$ as $\Phiv$ and to the Shapley values vector of $\x$ as $\phiv$.

\subsection{Counterfactual Explanations}\label{sec:counterfactual-explanations}

In its basic form, a \EAB{(local)} counterfactual explanation (CF) for an input $\x$ is a point $\xp$ such that (1) $\xp$ gives rise to a different prediction, i.e., $F(\x) \neq F(\xp)$, (2) $\x$ and $\xp$ are close (under some distance metric) and (3) $\xp$ is a ``plausible'' input.
This last constraint has been interpreted in several ways in the literature, it may involve considerations about sparsity \EAB{(e.g., \cite{Smyth}), proximity to the data manifold (e.g., \cite{Pawelczyk2020}), {proximity to other counterfactuals (e.g., \cite{Pawelczyk2021})}, causality (e.g., \cite{Karimi2021d}), actionability (e.g., \cite{Ustun,Poyiadzi2019}) or a combination thereof (e.g., \cite{Dandl2020})}.
A plethora of techniques for the generation of counterfactuals exist in the literature using search algorithms (e.g., \cite{Wachter2017,Albini2020,Albini2021Influence,Spooner2021}), optimization (e.g., \cite{Kanamori2020}) and genetic algorithms (e.g., \cite{Sharma2020}) among other methods (we refer the reader to \cite{Keane2021,Stepin2021,Karimi2020,Verma} for recent surveys).

In the scope of this paper we need to consider only counterfactual explanation methods that are (1) able to generate a set of (multiple) counterfactuals
and (2) do not require the model to be differentiable since 
we focus on tree-based models. We note that few counterfactual explanation techniques satisfying both of these requirements exist in the literature




\begin{figure*}[!ht]
\centering
\includegraphics[width=.99\textwidth]{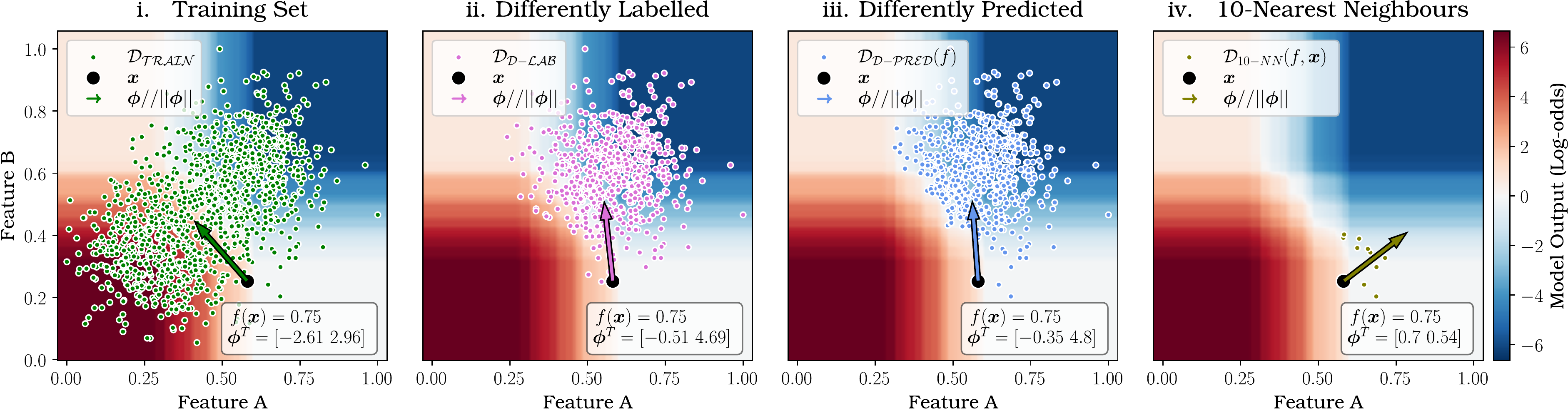}
\caption{
Effect of different choices of background dataset on the Shapley values ($\phiv$) of the same input ($x$) with the same model.
Red regions correspond to areas of the feature space where the decision is adverse, i.e. $F(\x) = 1$, with blue regions representing the opposite, i.e. those $\x \in \X$ for which $F(\x) = 0$.
Coloured arrows and scatter points represent the directions of the Shapley values vector and the background datasets used for their computation, respectively. \EB{We note how input-invariant distributions (i, ii, iii) do not give rise to SHAP values providing actionable guidance to overcome the adverse outcome; when using instead a set of counterfactuals (iv) a more actionable explanation is obtained (see Section~\ref{sec:choice-background-distribution} for more details).}
}
\label{fig:example_background}
\customlabel{fig:example_train}{\ref{fig:example_background}.i}
\customlabel{fig:example_dlab}{\ref{fig:example_background}.ii}
\customlabel{fig:example_dpred}{\ref{fig:example_background}.iii}
\customlabel{fig:example_knn}{\ref{fig:example_background}.iv}
\end{figure*}

\section{Counterfactual SHAP}

In general, Shapley values do not have an obvious interpretation in counterfactual terms, this means that they do not provide suggestions on how a user can change their features in order to change the prediction \cite{Kumar2020,Barocas2020}.
We argue that this is due to 2 main reasons: (1) the ``arbitrary'' choice of background distribution for the computation of Shapley values and \EB{(2) the lack of guidance on the best direction of the change for each of the features.}
\EB{We now discuss in details this two aspects.}


\subsection{Choice of the background distribution}\label{sec:choice-background-distribution}

Shapley values describe the contributions of the players (features) to the game payoff (model output). In the context of machine learning model explainability an important assumption is made: the simulation of each feature's 
absence in the cooperative game using a background distribution $\D$.
As pointed out in \cite{Merrick2020}, this means that 
Shapley values explain a prediction of an input \emph{\textbf{in contrast} to a distribution of background points}.
In practice, the background distribution is taken as a uniform distribution over unit point masses at a finite number of points, called the \emph{background dataset}.

Therefore, the background dataset should be chosen according to the contrastive question that \EE{one} aim to answer. We list some of the most common distributions that have been proposed.\footnote{\EE{We note that these distributions are often too large to use in practice, and instead the background dataset is obtained by sub-sampling or using k-means to generate a number of medoids \cite{Lundberg17,SHAPGithub2,SHAPGithub4}.}}
\begin{description}\setlength\itemsep{0em}
    \item[Training set] $\D_{\text{TRAIN}}$ \cite{Lundberg17,SHAPGithub2,SHAPGithub4}. The training set, including the samples that are labelled and/or predicted of being of the same class of the input.
    \item[Differently-labelled samples] $\D_{\text{D-LAB}}$ \cite{SHAPGithub1,SHAPGithub3}. The samples in the training set \emph{labelled} differently than the input.
    \item[Differently-predicted samples] \EB{$\D_{\text{D-PRED}}(f)$} \cite{SHAPGithub1,SHAPGithub3}. The samples in the training set \emph{predicted} with a different class.
\end{description}

These choices of background dataset have in common the fact that they are defined a priori, i.e., \EC{given a model}, they are equal for all the inputs.
This means that we are contrasting an input $\x$ with a (input-invariant) distribution $\D$ that may potentially be very different from $\x$.
This can give rise to explanations that are sometimes misleading for a user who is typically interested in understanding which features led to their adverse outcome (in order to reverse it) \cite{Miller2019}.
In other words the constrastive question that we are answering with the Shapley values is not tailored to the specific input (user) and therefore instead of answering the question of ``Why was {a user} rejected when compared to similar users that were accepted?'' we will be answering the more generic question of ``Which features are most important in making my outcome different from that of other (accepted) users?'' (who are potentially very different from $\x$).

The example in Figure~\ref{fig:example_train} shows an explanation where the background dataset is the training set, and we note that the Shapley values suggest that {Feature A} negatively contributed to the model output;
this means that the current value of {Feature A} is ``protective'' against rejection when \emph{put in contrast} with the expected output of the model obtained when using the background distribution $\D_{TRAIN}$. This may be useful information for the model developers \textbf{but it does not allow one to gain any (actionable) insight} unless we assume access to the underlying distribution $\D_{TRAIN}$.
In fact, this explanation only informs the user that their {Feature A} value has a positive impact when contrasted to typical {Feature A} values, but it does not either (a) advise them on how they can change their features in order to overcome the (adverse) outcome; or (b) inform them which features were most important in rejecting their application.

Figures~\ref{fig:example_dlab} and \ref{fig:example_dpred} 
show how alternative (but still  input-invariant) background distributions ($\D_{D-LAB}$ and $\D_{D-PRED}$, resp.) may improve the explanations in terms of informing the user on which features were most important in rejecting their application when compared to other rejected samples, but they still lack the ability of giving useful insight on which features were the most important and therefore should be acted upon in order to reverse the adverse decision $F(\x) = 1$.
This is due to the contrastive question being posed with respect to (a) samples that have much better (lower) model outputs 
and (b) samples that have similar model output but that are very different (in terms of distance in input space) from $\x$. 

Using a set of counterfactual points as the background dataset mitigates the issues mentioned in the preceding example.
In particular, using counterfactuals as the background dataset allows one to answer a contrastive question that is (a) of interest for the user because it is comparing $\x$ to samples that are similar to them (and implicitly more ``reachable'') and (b) more amenable in terms of access to the underlying distributions. In fact, as mentioned earlier, a useful interpretation of Shapley values-based explanations requires access to the background distribution.
Arguably, a user can relate to a set of similar customers more easily than the training set (that may contain very different users).
For example, it would be most useful in a credit lending context to compare customers with new accounts and no mortgage to other customers in a similar financial situation, while contrasting them to average customers, including many home-owners with older accounts, may lead to less actionable explanations.

\EB{We now formally define the Counterfactual SHAP values.

\begin{definition}\label{def:cshap_values}
    The \emph{Counterfactual SHAP values} for an input $\x \in \X$ are the Shapley values for $\x$ computed using the following characteristic function.
    $$
    v(S) = \E_{\xp \sim \D_C(f, \x)}\left[ f(\x_{S}, \xp_{\F \setminus S}) \right]
    $$
    where $C$ is (the name of) the counterfactual technique used to obtain the distribution $\D_C(f, \x)$.
\end{definition}
We note that, using a set of 
counterfactual points as a background distribution (contrary to ``classic'' input-invariant background distributions), means that the background distribution depends on the input $\x$ \EE{as reflected by Definition~\ref{def:cshap_values}}.
\EE{We can appreciate the effect of using counterfactuals as background dataset for the running example in Figure~\ref{fig:example_knn}. We note how both features are deemed as contributing to the rejection when compared to similar customers that were instead accepted. And in fact, Feature A has a higher importance than Feature B since the model is locally more sensitive to Feature A than Feature B as shown by the sharper color gradient in the horizontal direction.}


}

\begin{figure}[!t]
\centering
\includegraphics[height=.29\textwidth]{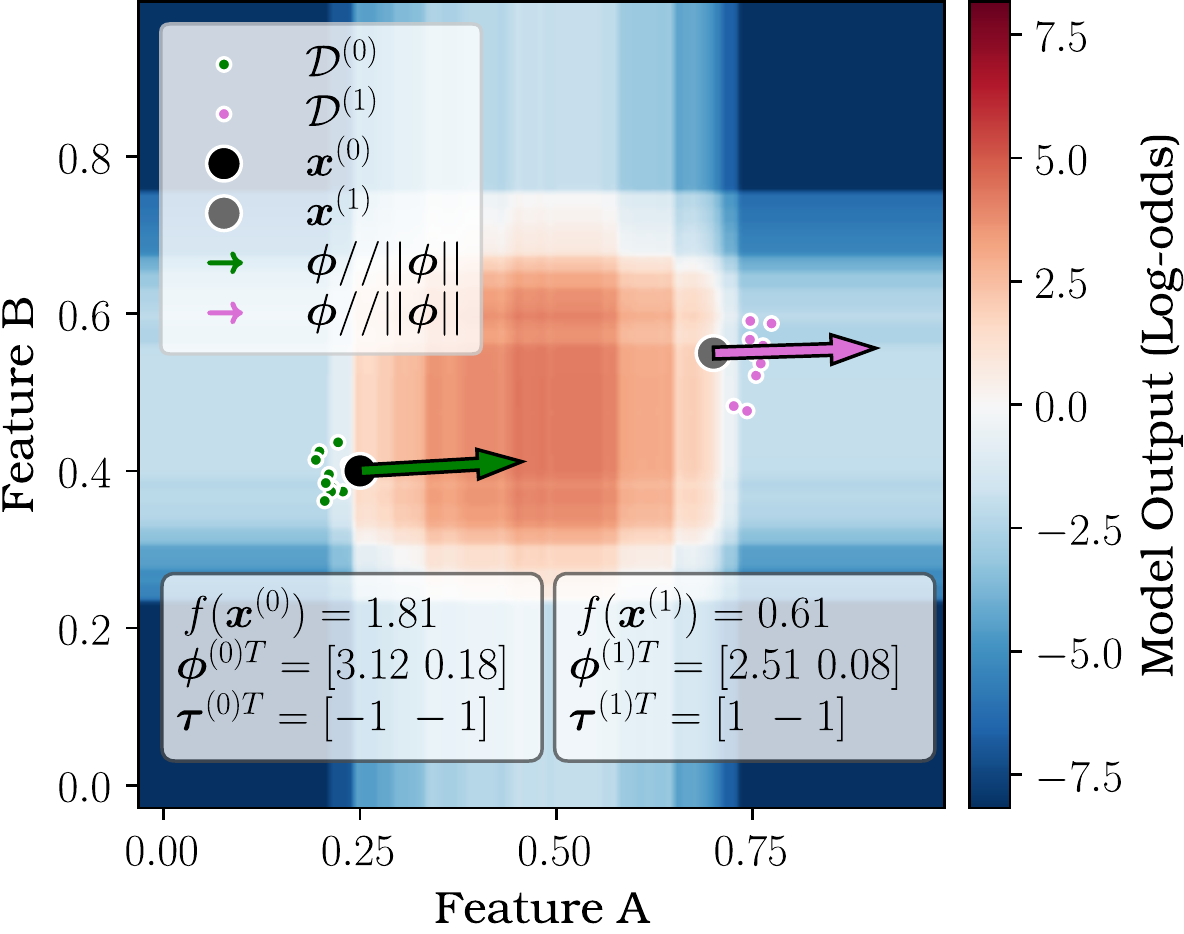}
\caption{Effect of (the lack of) \EB{derived trends} on the Shapley values of two same inputs using the $10$-NN as background dataset. 
Coloured arrows and scatter points represent the directions of the Shapley values vector and the background dataset used for their computation, respectively. We note that \EF{the derived trends better provide guidance for the direction of change of the features that cannot be afforded by the Shapley values alone} (see Section~\ref{sec:guidance_change} for more details).
}
\label{fig:examples1}
\customlabel{fig:example_trend}{\ref{fig:examples1}.i}
\end{figure}

\subsection{Guidance on the direction of the change}
\label{sec:guidance_change}

As remarked in \cite{Barocas2020}, feature attributions do not clearly provide guidance on how to alter the features in order to change the prediction of a model.
\EB{
In the case of Shapley values, this is due to Shapley values providing information about the most important features solely in terms of model output.
In practice this means that the Shapley value $\phi_i$ is not useful for understanding how one should alter the feature $i$ (i.e., increase or decrease it) in order to change the adverse outcome.
In other words, Shapley values identify features that have a strong impact on the output, \textbf{but not which inputs values are associated to output values of interest}. 
\EE{They are therefore not useful in informing a user on what intervention on $x_i$ is necessary to decrease $f(\x)$, and consequently overcome the adverse outcome.}

To address this issue we enrich the explanation provided by Counterfactual SHAP values with the \emph{derived trends}.
\begin{definition}
The \emph{derived trends} $\tauv \in \Tauv$ for an input $\x \in \X$ with respect to a distribution $\D_C(f, \x)$ are such that:
$$
\tau_i = sgn\left( \E_{\x' \in \D_{C}(f, \x)}\left[ x_i' \right] - x_i \right)
$$
where $sgn$ denotes the sign function\footnote{$sgn(x) = -1$ if $x<0$, $sgn(x) = +1$ if $x>0$, and $sgn(x) = 0$ otherwise.} and $\Tauv$ denotes the set of all derived trends (vectors), i.e., $\Tauv = \{ -1, +1, 0 \}^{\m}$.
\end{definition}
%
The derived trend $\tau_i$ for feature $i$ will be $+1$ ($-1$) when the value for the feature $\x_i$ is lower (higher, resp.) than that of the counterfactual distribution $\D_C(f, \x)$.
This means that, if $\tau_i = +1$ \EE{the user} must act upon $\x$ increasing $x_i$ to get closer to the expected value of $\D_C(f, \x)$, whilst if $\tau_i = -1$ \EE{they} must act upon $\x$ decreasing $x_i$ to achieve such goal.

To better understand the definition of the derived trends we can consider the example in Figure~\ref{fig:example_trend}, that shows two inputs $\x^{(0)}$ and $\x^{(1)}$ with similar Shapley values $\phiv^{(0)}$ and $\phiv^{(1)}$, respectively. 
It is evident that, despite {Feature A} being the most important feature for both inputs, in order to overcome the adverse outcome for $\x^{(0)}$ Feature A must be \emph{decreased} while it must be \emph{increased} for input $x^{(1)}$. 
This intuition is matched by the \EE{derived trends $\tau_{A}^{(0)} = -1$ and $\tau_{A}^{(1)} = +1$ correctly suggesting} that Feature A must be decreased or increased to change the decision for the input $x^{(0)}$ and $x^{(1)}$, respectively.
}

\subsection{Counterfactual SHAP}\label{sec:CFSHAP}

\EB{
After having described how the background distribution used for the computation of Shapley values plays a key role in giving a counterfactual interpretation to Shapley values and how \EE{explanations can be enriched with derived trends to provide guidance on direction of change of the features}, we now formally define Counterfactual SHAP explanations.

\begin{definition}
The \emph{Counterfactual SHAP} explanation (CF-SHAP) for an input $\x \in \X$ with respect to a distribution $\D_C(f, \x)$ is the tuple $(\phiv, \tauv) \in \Phiv \times \Tauv$ where $\phiv$ and $\tauv$ are, respectively, the \emph{Counterfactual SHAP values} and the \emph{derived trends} for $\x$ with respect to $\D_C(f, \x)$.
\end{definition}


We note that the various mathematical properties of Shapley values, and by extension SHAP values, have been studied in depth~\cite{Shapley1988,Lundberg17,Lundberg}. These properties typically fit with human intuition for feature importance and are used as part of the basis for justifying the SHAP framework. Since Counterfactual SHAP values apply the SHAP framework, Counterfactual SHAP also satisfies the key properties of \emph{additivity}, \emph{missingness} and \emph{consistency} as defined in~\cite{Lundberg17}. 
We provide a more thorough discussion in the supplementary material (see Appendix~\ref{appendix:algorithm}).

We now turn to the question of how we can numerically measure the ``counterfactual-ability'' of a feature attribution. 
We will tackle this problem in Section~\ref{sec:counterfactual-ability}.
}


\section{Counterfactual-ability}\label{sec:counterfactual-ability}






We seek to formalise the notion that certain feature attributions will be more useful for a model user in changing features to reverse an adverse outcome.
It is important to emphasise that predicting how users might engage with explanations is a very challenging problem, and behaviour may vary dramatically depending on the context. We do not claim to resolve this problem. 
However, we aim to set up a flexible framework to measure the ability of an explanation to help a user reverse an adverse decision (in Section~\ref{sec:counterfactual-ability-general}), before specialising this framework under certain \EB{sensible} assumptions about how a user could act on the explanations that they receive (in Section~\ref{sec:counterfactual-ability-instances}).


\subsection{General Evaluation Framework}\label{sec:counterfactual-ability-framework}\label{sec:counterfactual-ability-general}

To \EB{assess} 
the \emph{counterfactual-ability} of a feature attribution we \EB{first} use the explanation \EB{to generate what we call the \emph{induced counterfactual} point (from the explanation). 
Afterwards, we measure the ``goodness'' of the induced counterfactual point \EE{based on the cost that a user will incur when moving from the original input $\x$ to the induced counterfactual.} 
We expect a \EE{good feature attribution} to induce \EE{counterfactual points with lower cost and therefore higher counterfactual-ability}.
In our evaluation framework the \emph{induced counterfactual} from the explanation $(\phiv, \tauv)$ \EE{represents} the counterfactual point towards which the user will tend to move (under some sensible assumptions) with minimum cost (for the user). 

To formally define such notions we will now define two concepts: the \emph{cost function} and the \emph{action function}.}

\textbf{Cost Function}.
We measure the cost of changing an input $\x$ into another input $\x'$ via a \emph{cost function}. Formally, a \emph{{cost function}} is a function $\cf : \X \times \X \rightarrow \R^{\EA{+}}$ where $\cf(\x, \x')$ is the cost for the user of moving from $\x$ to $\x'$.
A \EE{trivial} example of cost function \EE{is} the Euclidean distance. 

\textbf{Action Function}.
In order to describe how a user acts upon the input $\x$ based on the information provided by the feature attribution $\phiv$ and the trends $\tauv$ we use an \emph{action function}.
Formally, an \emph{action function} is a function $\af : \X \times \Phiv \times \Tauv \rightarrow 2^{\X}$ where $\af(\x, \phiv, \tauv) \subset \X$ is a subset of the input space describing \EB{sensible} changes the user may enact upon $\x$ when provided with an explanation $(\phiv, \tauv)$. We will refer to $\af(\x, \phiv, \tauv)$ as the \emph{action subset}. Note that we do not constrain the action subset to be finite.

\EA{Intuitively the output of an action function can be interpreted as a subset of the possible options that a user may consider when changing the input based on the information provided by the \EB{explanation}.
For instance, a user may consider as possible options only changes to the most important feature according to the \EB{explanation}. In the most extreme scenario \EAB{a user may ignore the information provided by $\phiv$ and $\tauv$ and therefore} consider any change as a possible option; this would correspond to a constant action function always returning the whole input space as the action subset. 
\EAB{In a more realistic scenario though, we expect the user to use the information provided by the feature attribution and therefore we expect the action subset to be a restricted subset of the input space, e.g., allowing only changes to the top-3 most important features according to $\phiv$ \EB{and only in the directions suggested by $\tauv$}}.
}

\EB{
\textbf{Induced Counterfactual \EE{and Counterfactual-Ability}}.
\EE{The action and cost functions $\af$ and $\cf$ describe, respectively, (1) how a user may act upon $\x$ given an explanation and (2) how difficult it is for a user to perform such actions. Given $\af$ and $\cf$, the \emph{induced counterfactual} for an explanation
is then simply defined as the counterfactual point lying in the action subset such that a user has minimum cost to reach and the \emph{counterfactual-ability} is the negation of this cost.
We now formally define the notions of counterfactual-ability and induced counterfactual.} 
\begin{definition}
The \emph{counterfactual-ability} $\df(\x, \phiv, \tauv)$ of an explanation $(\phiv, \tauv)$ given an input $\x$ under an action function $\af$ and a cost function $\cf$ is defined as:
$$
\df(\x, \phiv, \tauv) = - \cf(\x, \x') \quad \text{where } \x' = \arginf_{\substack{\x' \in \af(\x, \phiv, \tauv)\\F(\x') \neq F(\x)}} \cf(\x, \x')
$$
and $\x' \in \X$ is referred to as the \emph{induced counterfactual} from $(\phiv, \tauv)$ given $\x$ under $\af$ and $\cf$.
\end{definition}
}

\EA{Note that the action function is fixed for a given user; the goal in fact is to compare how different feature attributions perform under a (given) action function rather than optimising the action function for a specific user.
We note that, in the degenerate case in which the action function is a constant function always returning the whole input space, solving this optimisation problem is equivalent to finding the counterfactual point $\x'$ with minimum cost from $\x$.
}

\begin{figure*}[!t]
\centering
\includegraphics[height=.29\textwidth]{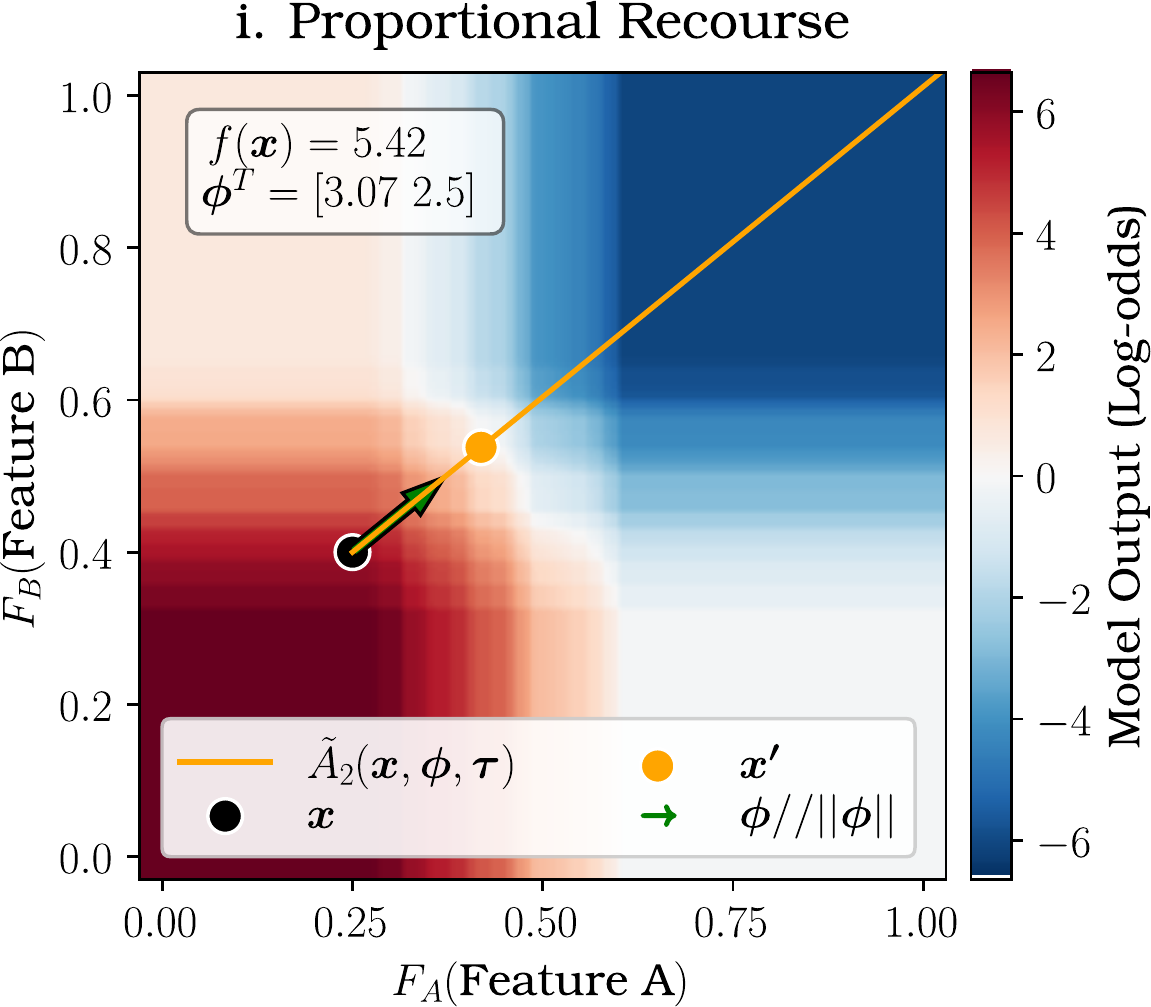}
\hspace{1cm}
\includegraphics[height=.29\textwidth]{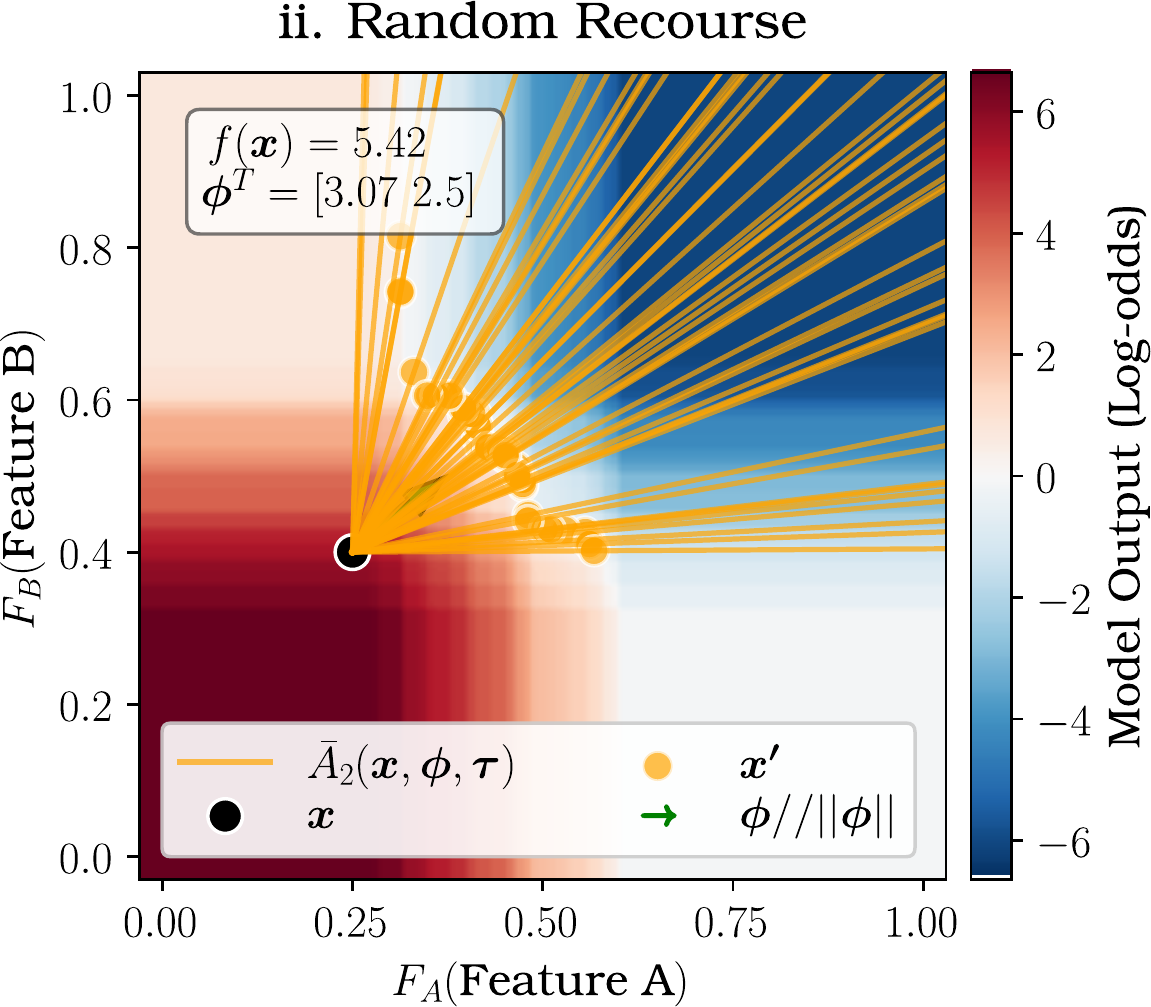}
\caption{
Algorithmic intuition of the action function. \EB{The orange lines in (i) and (ii) correspond to the action subsets $\tilde{A}_2(\x, \phiv, \tauv)$ (proportional) and $\bar{A}_2(\x, \phiv, \tauv)$ (random) as per definitions in Section~\ref{sec:counterfactual-ability-instances}. The corresponding orange points correspond to the points $\x'$} in the action subset with minimum cost according to L2-norm, \EE{i.e., the induced counterfactuals}. \EB{Note that the proportional action subset (line) in (i) has the same direction of the Shapley values vector while the \EF{random action subset in (ii) is uniformly} distributed.} Note that $F_A$ and $F_B$ are the cumulative distribution functions of Feature A and B, respectively. This means that the feature axes are in the quantile space as per definitions of the action functions.
}\label{fig:action_function}
\customlabel{fig:action_function_prop}{\ref{fig:action_function}.i}
\customlabel{fig:action_function_random}{\ref{fig:action_function}.ii}
\end{figure*}

Note that the \EB{more points are included in the action subset}, the smaller the counterfactual cost and therefore the larger the counterfactual-ability.
However, in order to realise the full potential of the counterfactual-ability a user must be able to find the minimal cost counterfactual within this subset, and this task is in general intractable for a user with limited access to the model. We must therefore make certain assumptions about the behaviour of a user. We will introduce the precise assumptions that we make in the following section, but our choices of action subsets will be restricted to {half-lines originating from the query point} in the input space. This choice emerges from the following informal assumption: a user will make an effort to change certain features in response to receiving an explanation, and the fraction of total effort that a user puts into changing each feature is not dependent on the total amount of effort that they expend. Assuming this, the action subset takes the form of a line in a direction determined by the manner in which the user chooses to apportion their effort in response to the explanation given.

\subsection{Instances of the Evaluation Framework}\label{sec:counterfactual-ability-instances}


After defining the general concepts of action function \EE{and} cost function 
we now define \EB{some} concrete instance\EB{s} that we use in this paper. 
\EB{To do so, we start with a number of assumptions designed to create sensible action and cost functions in the context of algorithmic recourse.}
Intuitively, the assumptions aim to cast the feature attribution as a suggested direction for a user to move in feature space, and the
\EE{counterfactual-ability}
will therefore measure \EE{the (negation of the) cost to reach the decision boundary along this line.}
In the scope of this paper, we \EA{use} the following set of assumptions:
\begin{description}\setlength\itemsep{0.em}
    \item[Assumption \customlabelnref{assumpt:trend}{1}: \EB{trend-aware} recourse.]
    When changing a feature a user moves \EA{its value} in the \EB{direction suggested by the \emph{derived trend} $\tau$, e.g., if $\tau_{income} = +1$, a user will try to increase their income (as opposed to reducing it) since this change is more likely to give rise to a change in the prediction.}
    \item[Assumption \customlabelnref{assumpt:adverse}{2}: adverse factors recourse.]
    A user changes \EE{only} features with \emph{positive} Shapley values, i.e, the features contributing to the adverse prediction (as opposed to also improving features that are already good).
    \item[Assumption \customlabelnref{assumpt:sparsity}{3}: sparse recourse.] 
    A user changes only the $k$ most important features (with the highest Shapley values). \EB{In many industry applications, e.g., the rejection of a loan application, regulations require companies to provide only the most important features \cite{EqualOpportunityAct} therefore making sensible to assume that users have access only to the most important features. The parameter $k$ will control how many features a user is allowed to change.}
    \item[Assumption \customlabelnref{assumpt:cost}{4}: recourse cost.]
    We use the \EE{\emph{quantile shift}} as metric to \EE{measure} the cost of the recourse, a common metric in the actionable recourse literature \cite{Ustun}. 
\end{description}

\begin{table*}[!t]
    \begin{small}
    \begin{center}
\begin{tabular}{r|ll}
\toprule
\textbf{Method} & \textbf{Distribution} & \textbf{Description} \\
\midrule
\multicolumn{2}{c}{\textsf{\textsc{Counterfactual SHAP}}} & \\
\midrule
CF-SHAP $K$-NN & $\D_{K\text{-NN}}(f,\x)$ & $K$-nearest neighbours of $\x$ such that $F(\x) = 1$ $^{\dagger}$\\
\midrule
\multicolumn{2}{c}{\textsf{\textsc{Baselines}}}\\
\midrule
 SHAP TRAIN  & $\D_{\text{TRAIN}}$ & Training set\\
 SHAP D-LAB  & $\D_{\text{DIFF-LAB}}$ & Samples in the training set such that $y = 1$ (label)\\
 SHAP D-PRED & $\D_{\text{DIFF-PRED}}(f)$ & Samples in the training set such that $F(\x) = 1$ (prediction)\\
\bottomrule
\end{tabular}


    \end{center}
    \protect\caption{
        \ED{Explanation methods used in the experiments divided among CF-SHAP and baselines. 
        ($\dagger$) We used $K=100$, see Appendix~\ref{appendix:increasingk} for additional experiments with a smaller or larger number of neighbours $K$.}
    }
    \label{table:methods}
    \end{small}
\end{table*}

Regarding the way in which a user do a recourse we have 2 \emph{alternative} assumptions that we will now introduce.

\begin{description}\setlength\itemsep{0.em}
    \item[Assumption \customlabelnref{assumpt:proportional}{5.A}: proportional recourse.] \EB{A user change the features \emph{proportionally} to their Shapley values: the higher the Shapley value the more a feature will be changed compared to others. In some applications users may be provided not only with the list of the most important features but also with the magnitude of the Shapley values for each of the features. The intuition behind this assumption is that some users may tend to change more some features (i.e., of a greater quantile shift) than others that are not deemed as important by the Shapley values.
    We will denote the following action function satisfying assumptions 1, 2, 3, 4 and 5.A with $\apropk$ where $k$ is the number of top features that a user considers.}\footnote{\label{footnote:odotprod}We use $\cdot$ and $\odot$ to indicate the dot and element-wise product, respectively.}
\end{description}
    \begin{equation}\label{eq:aprop}
    \begin{aligned}
        & \apropk(\x, \phiv, \tauv) = \\
        & \quad \big\{ \x' : \N(\x')-\N(\x) =  - \lambda \phiv \odot \tauv \odot \Ivec, \forall \lambda > 0 , \x' \in \X \big\}
    \end{aligned}
    \end{equation}
    where:
    \begin{itemize}[label={$\bullet$}]\setlength\itemsep{0.em}\setlength{\itemindent}{2em}
    \item 
    $\N : \X \rightarrow [0, 1]^{\m}$ is a function computing the quantile%
    \footnote{\label{footnote:quantile}%
    \EB{$\N : \X \rightarrow [0, 1]^{\m}$
    takes a vector $\x$ and returns a vector $\N(\x)$ with the marginal cumulative probability for each feature, 
    i.e., $\bm{q} = \N(\x)$ is such that $q_i = F_{i}(x_i)$ where $F_i$ is the cumulative distribution function of feature $i$ (estimated from the training set).}}
    of each of the features with respect to the \EB{marginal distributions in the training data};
    \item $\Ivec \in \{0, 1\}^{\m}$ is a (binary) indicator vector for the top-$k$ features in $\phiv$ \EB{with positive Shapley value}.\footnote{\EB{Formally the indicator vector $\Ivec$ for the top-$k$ features in $\phiv$ with positive Shapley values is such that for every element $\Iveci$ of $\Ivec$:
    $$
        \Iveci = \begin{cases}
            1 & \text{if } \phi_i > 0 \land i \in \argmax_{S \subseteq \{ 1, \ldots, \m \}, |S| \leq k}{\sum_{i \in S} \phi_i} \\
            0 & \text{otherwise}
        \end{cases}
    $$
    In other words, $\Iveci$ is $1$ iff $\phi_i$ is positive and $\phi_i$ is among the $k$ largest Shapley value in $\phi$, and $0$ otherwise.}
    }
    \end{itemize}

\EB{
The intuition behind this 
action function is that the feature attribution should provide a suggested direction to the user that takes them towards the decision boundary; we reflect this in Equation~\ref{eq:aprop} by using $\tauv$ to limit change in a certain direction (Assumption~\ref{assumpt:trend}) and by using $\phiv$ to enforce changes that are proportional to the feature attribution (Assumption~\ref{assumpt:proportional}).
However, realistic actions will not involve changes to every feature; rather, a user may focus on making changes to only the top-$k$ most important features (Assumption~\ref{assumpt:sparsity}) that are adversely contributing to the prediction (Assumption~\ref{assumpt:adverse}); we reflect this using $\Ivec$ in Equation~\ref{eq:aprop}.
We use the quantile shift as a normalised metric for recourse cost (Assumption~\ref{assumpt:cost}), enforced in Equation~\ref{eq:aprop} by normalizing $\x$ and $\x'$ with $\N$. Finally, note that $\lambda$ intuitively represents the amount of effort that a user put in changing their features, the higher $\lambda$ the farther $\x'$ is from $\x$.
}

The action subset induced by our action function is a semi-infinite line in the normalized quantile input space in the direction of the Shapley vector with its sign adjusted to match the monotonic trend. To better understand this concept we can consider Figure~\ref{fig:action_function_prop}, showing an example of the action \EA{subset} induced for an input $\x$ and an attribution $\phiv$.

\begin{description}\setlength\itemsep{0.em}  
    \item[Assumption \customlabelnref{assumpt:random}{5.B}: random recourse.] A user change the features \emph{randomly}. For every user a random ``utility vector'' will be used to describe which features should be changed more than others. The objective behind this assumption is two-fold; firstly, it introduces an element of robustness in the evaluation. In fact, we do not know how for different users some features may be more or less costly to change, and using a random utility among the top-$k$ features is an attempt to model this situation. Secondly, in some real-world applications users may not be provided with the Shapley values of each feature but only with a list of the top-$k$ most important features, therefore making the use of a proportional recourse infeasible. \EB{We will denote the following action function satisfying assumptions 1, 2, 3, 4 and 5.B with $\arandk$ where $k$ is the number of top features that a user considers.\textsuperscript{\ref{footnote:odotprod}}}
\end{description}

\begin{equation}\label{eq:arandom}
\begin{aligned}
    & \arandk(\x, \phiv, \tauv) = \\
    & \quad \big\{ \x' : \N(\x')-\N(\x) =  - \lambda \rv \odot \tauv \odot \Ivec, \forall \lambda > 0 , \x' \in \X \big\}
\end{aligned}
\end{equation}
where $\rv = (\rvar_1, \ldots, \rvar_{\m})$ is a vector of random variables following the distribution $\RD$. $\N$ and $\Ivec$ are defined as above.
\EC{In our experiments $\RD$ will be a uniform distribution between $0$ and $+1$.}





We note that our choice\EB{s} of action function \EB{are} just \EB{two} instantiation\EB{s} of the framework that we propose. 
\EC{We argue that casting the explanation as a \EC{random} direction in which an input point may move (Assumption~\ref{assumpt:random}) is a more robust choice than casting explanations in a proportional fashion (Assumption~\ref{assumpt:proportional}) as it makes fewer assumptions on the user preferences, but we acknowledge that there is no clear answer to the question of how different users may act upon $\x$ given an explanation $(\phiv, \tauv)$ in full generality.}


\textbf{Cost Function}.
\EF{We measured the cost using two alternative definitions based on the features quantile shift: the quantile shift under L1-norm (a.k.a., total quantile shift \cite{Ustun}) and L2-norm.}
Formally\textsuperscript{\ref{footnote:quantile}}:
$$
\cf_{L1}(\x, \x') =  \norm{\N(\x') - \N(\x)}_1 \spaceornewline{,\ }
\cf_{L2}(\x, \x') =  \norm{\N(\x') - \N(\x)}_2.
$$

\EC{We note that the instances of action and cost functions that we propose implicitly assume that the changes to features are made through interventions. 
We acknowledge that in some applications in which a full causal understanding of the input space is possible, one could instead resort to changes that take into consideration the causal structure.
We believe that this topic represents an interesting future research direction \EC{(see Section~\ref{sec:conclusions})}.} 

\begin{figure*}[!t]
\centering
\includegraphics[width=.95\textwidth]{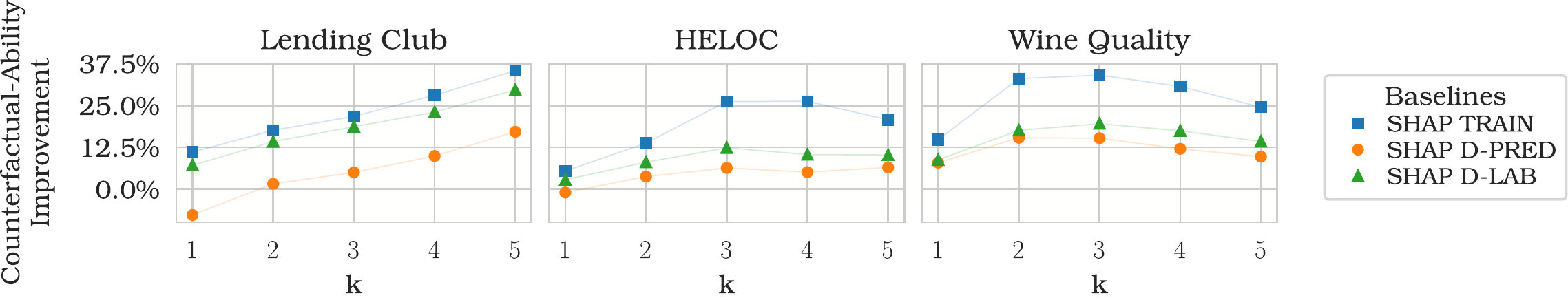}
\caption{
\ED{Counterfactual-ability improvement (as defined in Section~\ref{sec:experiments}) of CF-SHAP with respect to the baselines}
(SHAP TRAIN, SHAP D-LAB,SHAP D-PRED). The plots show how the \EB{counterfactual-ability improves} when varying the number of top-$k$ features \ED{a user has access to}.  In particular, this plots show the results for the improvement in counterfactual-ability under $\cf_{L1}$ cost function and random recourse ($\arandk$). Each line represents a baseline.
}\label{fig:exp_costsL1}
\end{figure*}

\section{Experiments}\label{sec:experiments}





In order to understand how \EB{Counterfactual SHAP} (CF-SHAP) performs we compared it against existing feature attribution techniques. 
\ED{In particular, we compared CF-SHAP with SHAP using common input-invariant background distributions}: $\D_{TRAIN}, \D_{D-PRED}$ and $\D_{D-LAB}$. Table \ref{table:methods} \EB{summarizes the \ED{baselines}} that we considered in our experiments. We refer to Section~\ref{sec:choice-background-distribution} for more details about these distributions. 


\ED{To generate counterfactual points for Counterfactual SHAP we used $K$-nearest neighbours ($K$-NN). In practice,} 
to generate counterfactual points for $\x$ 
we took the $K$ nearest points to $\x$ in the training set, such that their predictions were different from the prediction for $\x$. In our experiments we used $K=100$ and the Manhattan distance over the quantile space as distance metric for neighbours. This distribution will be referred to as $\D_{K-NN}(\x, f)$. \ED{We note that we set $K=100$ when running our experiments in order to allow for a fair comparison with SHAP that, by default, randomly samples $100$ points from the background distribution that it has been given.}

\ED{
We choose $K$-NN as technique rather than more complex counterfactual generation engines because (1) few counterfactual generation techniques are able to generate multiple counterfactuals and even fewer a diverse set thereof, especially in the context of decision tree-based models; (2) most importantly, the choice of $K$-NN as the technique for the generation of counterfactuals allows us to showcase the performance of Counterfactual SHAP while separating it from the performance of the counterfactual generation engine used. For example, using $K$-NN allows us to generate counterfactual points that are on-manifold (since they are points of the training set), an issue with which many counterfactual generation techniques struggle \cite{Keane2021}.
}

To run the experiments we used \ED{3} publicly available datasets: 
\textbf{HELOC} (Home Equity Line Of Credit) \cite{FICO}, 
\textbf{LC} (Lending Club Loan Data) \cite{LendingClub} and 
\textbf{WQ} (UCI Wines Quality) \cite{Cortez2009}.
\EC{For each dataset,} we trained an XGBoost model \cite{XGBoost} \ED{and used TreeSHAP \cite{Lundberg2020Trees} to generate explanations}.
\EB{We refer to Appendix~\ref{appendix:setup} for more details about the datasets and the experimental setup.}

\begin{figure*}[!t]
\centering
\includegraphics[width=.95\textwidth]{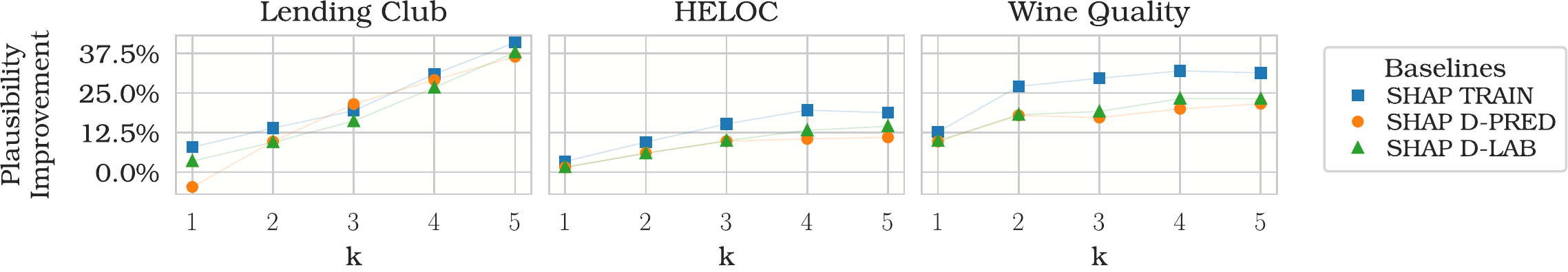}
\caption{
\ED{Plausibility improvement (as defined in Section~\ref{sec:experiments}) of CF-SHAP with respect to the baselines}
(SHAP TRAIN, SHAP D-LAB,SHAP D-PRED). The plots show how the \EB{plausibility improves} when varying the number of top-$k$ features \ED{a user has access to}. In particular, this plots show the results for the improvement in plausibility under L1-norm ($\pf_{L1}$), $\cf_{L1}$ cost function and random recourse ($\arandk$). Each line represents a baseline.
}\label{fig:exp_plausaL1}
\end{figure*}

\EB{\EE{\textbf{Counterfactual-ability.}}}
\EF{Counterfactual-ability represents a key metric to evaluate explanations in counterfactual terms as it measures the ability of a feature attribution to help a user reverse an adverse decision with minimal cost. We expect good feature attributions to have higher counterfactual-ability.}
\EC{For each dataset,} we measured the percentage of times in which CF-SHAP \EB{has higher counterfactual-ability} than that of the baselines
and subtracted the number of times in which CF-SHAP has instead lower counterfactual-ability.
We measured this over \ED{$4,000$} rejected (i.e. with $F(\x) = 1$) random samples in the test set \ED{(or all of the rejected samples if less than $4,000$ were available)}.
Formally, we measure the improvement in counterfactual-ability as follows.

\begin{widetext}
$$
\parbox{10em}{%
\raggedleft Counterfactual-Ability\tiny\\\normalsize Improvement}\ = 
    \E_{x \in D_{F(\x) = 1}}\left[\Indicator{
        \df(\x, \phiv, \tauv) > \df(\x, \phiv_{BASE}, \tauv^{G})
    }
    -
     \Indicator{
        \df(\x, \phiv, \tauv) < \df(\x, \phiv_{BASE}, \tauv^{G})
    }\right]
$$
\end{widetext}

where $(\phiv, \tauv)$ and $(\phiv_{BASE}, \tauv^{G})$ are the the explanations for CF-SHAP and the baseline, respectively, $D_{F(\x) = 1}$ is the set of rejected samples and $\BigOne$ is the boolean indicator\footnote{$\Indicator{\cdot} = 1$ if the boolean expression $\cdot$ is true, and $\Indicator{\cdot} = 0$ otherwise.}.    
Figure~\ref{fig:exp_costsL1} 
show the results using the \EF{random \EC{action function} ($\apropk$) and the cost function with L1-norm ($\cf_{L1}$). We refer to the supplementary material (see Appendix~\ref{appendix:moreexperiments}) for additional results with alternative definitions of action function and cost function.}
We report the main findings.

\ED{
\begin{itemize}\setlength\itemsep{0em}
    \item Despite $K$-NN being a crude method for counterfactual generation, CF-SHAP beats (i.e., $>0\%$) the baselines for all 3 datasets.
    \item \EF{There seems to be a positive correlation between the number of top-$k$ features that are allowed to change and the improvement in counterfactual-ability. This suggests that CF-SHAP is able not only to find the top-1 most important feature to change to reverse the prediction, but it is also better performing than the baselines in identifying the top-2 and top-3 most important features that a user must change to reverse an adverse outcome.}
    \item The results are robust with respect to the action function ($\arandk$ or $\apropk$) and cost function ($\cf_{L1}$, $\cf_{L2}$) used. We refer to Appendix~\ref{appendix:moreexperiments} for details on this results.
    \item Depending on the dataset the best choice for the hyper-parameter $K$ of $K$-NN ranges from $K=10$ to $K=100$. We refer to Appendix~\ref{appendix:increasingk} for detailed experiments on this hyper-parameter.
\end{itemize}
}

\EF{
We note that to allow for a fair comparison of the counterfactual-ability of CF-SHAP with that of the baselines
(1) we equipped explanations
that do not provide a derived trend with a ``global'' trend $\tauv^{G}$ obtained using the Pearson correlation of the features and target values in the training set\footnote{Formally, given the training set $\left[ \bm{X}_{1}, \ldots, \bm{X}_{m} \right], \bm{y}$, we define the \emph{global trend} as a vector $\tauv^{G} \in \{-1,0+1\}^{\m}$ such that:
$$
\tau^{G}_i = sgn\left( \rho_{\bm{X}_{i}, \bm{y}} \right) \quad \forall i \in \{1, \ldots, \m\}
$$
where $sgn$ is the sign function and $\rho_{\bm{X}_{i}, \bm{y}}$ is the Pearson's r between the column $\bm{X}_{i}$ of the training set (corresponding to feature $i$) and the labels vector $\bm{y}$.
};
(2) in order to to implement the random action function, we generated a random vector $\rv$ (see Equation~\ref{eq:arandom}) for each sample. This means that, as one would expect, all explanation \EF{techniques} have been tested using the same ``utility vectors'' (one for each sample) modelling different users' preferences on the features to change;
(3) the counterfactual-ability results are reported in relative terms to allow for the comparison of explanations for which counterfactual-ability is $-\infty$ (when no induced counterfactual can be found in the action set).
}

\EB{\textbf{Plausibility.}} \EE{As pointed out in Section~\ref{sec:counterfactual-explanations}, plausibility is another important desideratum for counterfactuals. 
We expect a good feature attribution to induce counterfactuals that are not less plausible than those that can be found using input-invariant distributions. 
We measured the \emph{plausibility} of the induced counterfactual in terms of the density of the region in which it lies so that, in practice, we computed the average distance of the induced counterfactual from its $5$ nearest neighbours (measured again using quantile shifts).
Formally, we measure plausibility in two ways:
$$
    \pf_{L1}(\x, \x') =  - \E_{\x^* \in \NN(\x', 5)}\left[\norm{\N(\x^*) - \N(\x')}\right]_1 \spaceornewline{, \ \ }
    \pf_{L2}(\x, \x') =  - \E_{\x^* \in \NN(\x', 5)}\left[\norm{\N(\x^*) - \N(\x')}\right]_2 
$$
where $\NN$ is a function $\NN : \X \times \NNZ \rightarrow 2^{\X}$ such that $\NN(\x', n)$ is the set of the $n$ nearest neighbours of $\x'$.
Similarly to the counterfactual-ability, we computed the percentage of times in which CF-SHAP \EB{has higher plausibility} than that of the baselines \ED{and subtracted the number of times in which CF-SHAP has lower counterfactual-ability, i.e., formally:
}

\begin{widetext}
$$
\parbox{7em}{%
\raggedleft Plausibility\tiny\\\normalsize Improvement}\ = 
    \E_{x \in D_{F(\x) = 1}}\left[\Indicator{
        \pf(\x, \x_{CF-SHAP}') > \pf(\x, \x_{BASE}')
    }
    -
     \Indicator{
        \pf(\x, \x_{CF-SHAP}') < \pf(\x, \x_{BASE}')
    }\right]
$$
\end{widetext}

where $\x_{CF-SHAP}$ and $\x_{BASE}$ are the the induced counterfactuals for CF-SHAP and the baseline, respectively, and $D_{F(\x) = 1}$ is the set of rejected samples.
}
We run the experiments over $4,000$ rejected (i.e. with $F(\x) = 1$) random samples for each dataset.
Figure~\ref{fig:exp_plausaL1} 
shows the results using random recourse (i.e., action function $\arandk$) and total quantile shift cost (i.e., cost function $\cf_{L1}$). 
We note that CF-SHAP beats (i.e., $>0\%$) the baselines in plausibility for all datasets and it is highly robust to the choice of plausibility normalisation, action function and cost function. We refer to Appendix~\ref{appendix:moreexperiments} for detailed results with alternative action function, cost function and plausibility normalisation.

\EB{
\textbf{Execution Time}.
The execution time of CF-SHAP directly depends on the execution time of (1) the counterfactual generation technique and (2) the execution time of SHAP. 
We note the computational complexity of (Tree-)SHAP scales linearly with the size of the background dataset \cite{Lundberg17}.
\ED{Our experiments showed that CF-SHAP has a similar execution time to the baselines. This means that $K$-NN do not add a significant overhead to the overall execution time of CF-SHAP. In particular, CF-SHAP explanations could be generated (on average) \EF{in as little as $620\mu s$ for the Wine Quality dataset, $646\mu s$ for the HELOC dataset and $2646 \mu s$ for the Lending Club dataset}.}
We refer to the supplementary material (see Appendix~\ref{appendix:performance}) for a detailed benchmark on the execution time of CF-SHAP.
}

\section{Related Work}\label{sec:related}

There has been recently an increasing interest in exploring the relationship between feature importance and counterfactual explanations. 
\EA{A recent work \cite{Watson2021} has proposed a Bayesian decision theory-based approach to the computation of the Shapley values. In particular the idea of \cite{Watson2021} is to optimize the choice of the background distribution for the computation of Shapley values maximizing the expected reward for the user, i.e., $\D^{*} = \argmax_{\D^{*} \subseteq \D} E_{\x \sim \D^{*}}[r(\x)]$, under a certain reward function $r$. 
The work provides a theoretical framework for modelling user preference and beliefs but lacks (by design) concrete (1) guidance on how to select $\D^{*}$, (2) how to update the reward function $r$ based on the observed Shapley values and (3) how to interpret the feature attribution $\phiv$ into practical actions on the input $\x$ in order to (automatically) solve the optimisation problem without resorting to an update in human-in-the-loop fashion.
}

Other works have proposed to fill the gap between feature attribution techniques and counterfactual explanations by different means than Shapley values. 
In particular, \cite{Mothilal2021} and \cite{Sharma2020} propose techniques to generate feature attributions \ED{for differentiable models} from a set of diverse counterfactual points but (contrary to us) they use frequency-based approaches, i.e., they give higher attribution to features that are more often changed in counterfactual points. This implies that also features \EA{potentially ignored by the model may receive a high feature importance because they are correlated with other features that are really used by the model. As remarked in \cite{Chen} this behaviour may be desirable in some context as medical sciences but not in others, as in the credit scoring scenario in which users are ultimately interested in understanding why they have have been rejected \emph{by the model} rather than which features correlate with rejection \emph{in the data}}. 
In \cite{Chapman-Rounds2021} feature attributions are generated by approximating the minimal adversarial perturbation using an adversarially trained neural network on a (differentiable) neural network-based surrogate model. This approach tends to follow the most strictest interpretation of the ``true to the model'' paradigm \cite{Chen} enforcing only the class change but does not directly allow for the enforcement of other constraints, e.g., regarding the plausibility of such changes, as we do by providing a background distribution that is based on counterfactuals.

Other works \cite{Ramon2020,Rathi2019,Fernandez-Loria2021,Fern2021} analyze the complementary problem to that we analyze in this paper: they show how feature attributions can guide the search of counterfactuals (while we investigate how techniques for the generation of counterfactuals can empower better feature attribution).

In general, many works have explored how to evaluate counterfactual explanations (e.g., \cite{Ustun,Rawal2020,Laugel2019}) and feature attributions (e.g., \cite{Guidotti2018LORE,Plumb2018,Lakkaraju2019}) but few proposed a quantitative metric to evaluate feature attributions in counterfactual terms.
In \cite{White2019} the authors propose to evaluate feature attributions with a \emph{fidelity error} for each of the features that (differently from counterfactual-ability) can be computed changing only a single feature at a time. \ED{We overcome this limitation with the parameter $k$, controlling the number of features that are allowed to change.}

\section{Conclusion and future work}\label{sec:conclusion}\label{sec:conclusions}



Towards the more general goal of unifying feature attribution techniques and counterfactual explanations, we have shown how using \ED{a set of counterfactuals} as the background distribution for the computation of Shapley values allows one to obtain feature attributions that can better advise towards useful changes of the input in order to overcome an adverse outcome. \ED{We have also highlighted that the generation of feature attributions with a counterfactual intent requires one to enrich explanations with additional information to describe the direction of the change (i.e., the derived trends).}
We proposed a new quantitative framework to evaluate such an effect \ED{in terms of counterfactual-ability and plausibility based on the notions of action and cost functions}.
We evaluated CF-SHAP on 3 publicly available datasets and highlighted that using simpler counterfactual techniques such as those based on nearest-neighbours within CF-SHAP performs better than existing feature attribution methods.

Our proposal can be extended in several directions.
Firstly, it would be interesting to explore alternative notions of action and cost functions, grounding their definition with findings in psychology concerning how users interpret feature attributions and how they consequently change their behaviour.
For example, one possibility would be to expand the definition of action function to take into account user preferences for certain actions -- this could be achieved by coupling each ``possible action'' returned by the action function with a probability.
\EB{Secondly}, testing our approach on different models (e.g., neural networks) and using (potentially model-agnostic) counterfactual explanation techniques as \cite{Karimi2020,Karimi2021d,Dandl2020, Kanamori2020,Rawal2020,Hashemi2020} represents another interesting future direction.
\EC{Lastly, investigating how the generation of feature attribution with a counterfactual flavour connects with causality would also be of great interest. This could be achieved by considering a conditional background distribution in the computation of Shapley values \EC{\cite{Janzing2020,Aas2021,Frye2021,Wang2021,Sundararajan2020}} or using counterfactual explanation techniques that explicitly make use of causality, e.g. \EC{\cite{vonKugelgen2022, Karimi2020a}}.}

From a wider perspective, our work draws attention to some gaps in the literature that we believe are worthy of further investigation.
On the one hand, the importance of techniques for the generation of \EC{diverse} counterfactuals advocated by many practitioners \cite{Mothilal2020a, Russell2019, Smyth}. On the other hand, it highlights how few techniques have the capabilities of \EB{efficiently} generating diverse counterfactual explanations in the context of non-differentiable models that are among the most widely adopted in industry, e.g., ensembles of decision trees.


\begin{acks}
\textbf{Disclaimer}. 
This paper was prepared for informational purposes by
the Artificial Intelligence Research group of JPMorgan Chase \& Co. and its affiliates (``JP Morgan''),
and is not a product of the Research Department of JP Morgan.
JP Morgan makes no representation and warranty whatsoever and disclaims all liability,
for the completeness, accuracy or reliability of the information contained herein.
This document is not intended as investment research or investment advice, or a recommendation,
offer or solicitation for the purchase or sale of any security, financial instrument, financial product or service,
or to be used in any way for evaluating the merits of participating in any transaction,
and shall not constitute a solicitation under any jurisdiction or to any person,
if such solicitation under such jurisdiction or to such person would be unlawful.
\end{acks}

\bibliographystyle{ACM-Reference-Format}
\bibliography{main}


\begin{thebibliography}{70}


\ifx \showCODEN    \undefined \def \showCODEN     #1{\unskip}     \fi
\ifx \showDOI      \undefined \def \showDOI       #1{#1}\fi
\ifx \showISBNx    \undefined \def \showISBNx     #1{\unskip}     \fi
\ifx \showISBNxiii \undefined \def \showISBNxiii  #1{\unskip}     \fi
\ifx \showISSN     \undefined \def \showISSN      #1{\unskip}     \fi
\ifx \showLCCN     \undefined \def \showLCCN      #1{\unskip}     \fi
\ifx \shownote     \undefined \def \shownote      #1{#1}          \fi
\ifx \showarticletitle \undefined \def \showarticletitle #1{#1}   \fi
\ifx \showURL      \undefined \def \showURL       {\relax}        \fi
\providecommand\bibfield[2]{#2}
\providecommand\bibinfo[2]{#2}
\providecommand\natexlab[1]{#1}
\providecommand\showeprint[2][]{arXiv:#2}

\bibitem[\protect\citeauthoryear{Aas, Jullum, and L{\o}land}{Aas
  et~al\mbox{.}}{2021}]%
        {Aas2021}
\bibfield{author}{\bibinfo{person}{Kjersti Aas}, \bibinfo{person}{Martin
  Jullum}, {and} \bibinfo{person}{Anders L{\o}land}.}
  \bibinfo{year}{2021}\natexlab{}.
\newblock \showarticletitle{Explaining Individual Predictions When Features Are
  Dependent: {{More}} Accurate Approximations to {{Shapley}} Values}.
\newblock \bibinfo{journal}{\emph{Artificial Intelligence}}
  \bibinfo{volume}{298} (\bibinfo{year}{2021}), \bibinfo{pages}{103502}.
\newblock


\bibitem[\protect\citeauthoryear{Adadi and Berrada}{Adadi and Berrada}{2018}]%
        {Adadi2018}
\bibfield{author}{\bibinfo{person}{Amina Adadi} {and} \bibinfo{person}{Mohammed
  Berrada}.} \bibinfo{year}{2018}\natexlab{}.
\newblock \showarticletitle{{Peeking Inside the Black-Box: A Survey on
  Explainable Artificial Intelligence (XAI)}}.
\newblock \bibinfo{journal}{\emph{IEEE Access}}  \bibinfo{volume}{6}
  (\bibinfo{date}{9} \bibinfo{year}{2018}), \bibinfo{pages}{52138--52160}.
\newblock


\bibitem[\protect\citeauthoryear{Albini, Rago, Baroni, and Toni}{Albini
  et~al\mbox{.}}{2020}]%
        {Albini2020}
\bibfield{author}{\bibinfo{person}{Emanuele Albini}, \bibinfo{person}{Antonio
  Rago}, \bibinfo{person}{Pietro Baroni}, {and} \bibinfo{person}{Francesca
  Toni}.} \bibinfo{year}{2020}\natexlab{}.
\newblock \showarticletitle{Relation-{{Based Counterfactual Explanations}} for
  {{Bayesian Network Classifiers}}}. In \bibinfo{booktitle}{\emph{Proceedings
  of the 29th International Joint Conference on {{Artificial Intelligence}},
  IJCAI}}. \bibinfo{pages}{451--457}.
\newblock


\bibitem[\protect\citeauthoryear{Albini, Rago, Baroni, and Toni}{Albini
  et~al\mbox{.}}{2021}]%
        {Albini2021Influence}
\bibfield{author}{\bibinfo{person}{Emanuele Albini}, \bibinfo{person}{Antonio
  Rago}, \bibinfo{person}{Pietro Baroni}, {and} \bibinfo{person}{Francesca
  Toni}.} \bibinfo{year}{2021}\natexlab{}.
\newblock \showarticletitle{Influence-Driven Explanations for Bayesian Network
  Classifiers}. In \bibinfo{booktitle}{\emph{PRICAI 2021: Trends in Artificial
  Intelligence}}. \bibinfo{pages}{88--100}.
\newblock


\bibitem[\protect\citeauthoryear{Barocas, Selbst, and Raghavan}{Barocas
  et~al\mbox{.}}{2020}]%
        {Barocas2020}
\bibfield{author}{\bibinfo{person}{Solon Barocas}, \bibinfo{person}{Andrew~D
  Selbst}, {and} \bibinfo{person}{Manish Raghavan}.}
  \bibinfo{year}{2020}\natexlab{}.
\newblock \showarticletitle{{The Hidden Assumptions Behind Counterfactual
  Explanations and Principal Reasons}}. In
  \bibinfo{booktitle}{\emph{Proceedings of the 2020 Conference on Fairness,
  Accountability, and Transparency, FAccT}}. \bibinfo{pages}{80--89}.
\newblock


\bibitem[\protect\citeauthoryear{Barredo~Arrieta, D{\'{i}}az-Rodr{\'{i}}guez,
  Del~Ser, Bennetot, Tabik, Barbado, Garcia, Gil-Lopez, Molina, Benjamins,
  Chatila, and Herrera}{Barredo~Arrieta et~al\mbox{.}}{2020}]%
        {BarredoArrieta2020}
\bibfield{author}{\bibinfo{person}{Alejandro Barredo~Arrieta},
  \bibinfo{person}{Natalia D{\'{i}}az-Rodr{\'{i}}guez}, \bibinfo{person}{Javier
  Del~Ser}, \bibinfo{person}{Adrien Bennetot}, \bibinfo{person}{Siham Tabik},
  \bibinfo{person}{Alberto Barbado}, \bibinfo{person}{Salvador Garcia},
  \bibinfo{person}{Sergio Gil-Lopez}, \bibinfo{person}{Daniel Molina},
  \bibinfo{person}{Richard Benjamins}, \bibinfo{person}{Raja Chatila}, {and}
  \bibinfo{person}{Francisco Herrera}.} \bibinfo{year}{2020}\natexlab{}.
\newblock \showarticletitle{{Explainable Explainable Artificial Intelligence
  (XAI): Concepts, taxonomies, opportunities and challenges toward responsible
  AI}}.
\newblock \bibinfo{journal}{\emph{Information Fusion}}  \bibinfo{volume}{58}
  (\bibinfo{year}{2020}), \bibinfo{pages}{82--115}.
\newblock


\bibitem[\protect\citeauthoryear{Bergstra, Yamins, and Cox}{Bergstra
  et~al\mbox{.}}{2013}]%
        {hyperopt}
\bibfield{author}{\bibinfo{person}{James Bergstra}, \bibinfo{person}{Daniel
  Yamins}, {and} \bibinfo{person}{David Cox}.} \bibinfo{year}{2013}\natexlab{}.
\newblock \showarticletitle{Making a Science of Model Search: Hyperparameter
  Optimization in Hundreds of Dimensions for Vision Architectures}. In
  \bibinfo{booktitle}{\emph{Proceedings of the 30th International Conference on
  International Conference on Machine Learning, ICML}}.
  \bibinfo{pages}{I–115–I–123}.
\newblock


\bibitem[\protect\citeauthoryear{Chapman-Rounds, Bhatt, Pazos, Schulz, and
  Georgatzis}{Chapman-Rounds et~al\mbox{.}}{2021}]%
        {Chapman-Rounds2021}
\bibfield{author}{\bibinfo{person}{Matt Chapman-Rounds}, \bibinfo{person}{Umang
  Bhatt}, \bibinfo{person}{Erik Pazos}, \bibinfo{person}{Marc-Andre Schulz},
  {and} \bibinfo{person}{Konstantinos Georgatzis}.}
  \bibinfo{year}{2021}\natexlab{}.
\newblock \showarticletitle{{FIMAP: Feature Importance by Minimal Adversarial
  Perturbation}}.
\newblock \bibinfo{journal}{\emph{Proceedings of the 35th AAAI Conference on
  Artificial Intelligence}} \bibinfo{volume}{35}, \bibinfo{number}{13}
  (\bibinfo{date}{5} \bibinfo{year}{2021}), \bibinfo{pages}{11433--11441}.
\newblock


\bibitem[\protect\citeauthoryear{Chen, Janizek, Lundberg, and Lee}{Chen
  et~al\mbox{.}}{2020}]%
        {Chen}
\bibfield{author}{\bibinfo{person}{Hugh Chen}, \bibinfo{person}{Joseph~D
  Janizek}, \bibinfo{person}{Scott Lundberg}, {and} \bibinfo{person}{Su-In
  Lee}.} \bibinfo{year}{2020}\natexlab{}.
\newblock \showarticletitle{{True to the Model or True to the Data?}}. In
  \bibinfo{booktitle}{\emph{ICML '20 Workshop on Human Interpretability}}.
\newblock
\showeprint[arxiv]{2006.16234}


\bibitem[\protect\citeauthoryear{Chen and Guestrin}{Chen and Guestrin}{2016}]%
        {XGBoost}
\bibfield{author}{\bibinfo{person}{Tianqi Chen} {and} \bibinfo{person}{Carlos
  Guestrin}.} \bibinfo{year}{2016}\natexlab{}.
\newblock \showarticletitle{{XGBoost}: A Scalable Tree Boosting System}. In
  \bibinfo{booktitle}{\emph{Proceedings of the 22nd ACM SIGKDD International
  Conference on Knowledge Discovery and Data Mining, KDD}}.
  \bibinfo{pages}{785--794}.
\newblock


\bibitem[\protect\citeauthoryear{Cortez, Cerdeira, Almeida, Matos, and
  Reis}{Cortez et~al\mbox{.}}{2009}]%
        {Cortez2009}
\bibfield{author}{\bibinfo{person}{Paulo Cortez}, \bibinfo{person}{António
  Cerdeira}, \bibinfo{person}{Fernando Almeida}, \bibinfo{person}{Telmo Matos},
  {and} \bibinfo{person}{José Reis}.} \bibinfo{year}{2009}\natexlab{}.
\newblock \showarticletitle{{Modeling wine preferences by data mining from
  physicochemical properties}}.
\newblock \bibinfo{journal}{\emph{Decision Support Systems}}
  \bibinfo{volume}{47}, \bibinfo{number}{4} (\bibinfo{date}{11}
  \bibinfo{year}{2009}), \bibinfo{pages}{547--553}.
\newblock


\bibitem[\protect\citeauthoryear{{\v C}yras, Rago, Albini, Baroni, and
  Toni}{{\v C}yras et~al\mbox{.}}{2021}]%
        {Cyras2021}
\bibfield{author}{\bibinfo{person}{Kristijonas {\v C}yras},
  \bibinfo{person}{Antonio Rago}, \bibinfo{person}{Emanuele Albini},
  \bibinfo{person}{Pietro Baroni}, {and} \bibinfo{person}{Francesca Toni}.}
  \bibinfo{year}{2021}\natexlab{}.
\newblock \showarticletitle{Argumentative {{XAI}}: A {{Survey}}}. In
  \bibinfo{booktitle}{\emph{Proceedings of the 29th International Joint
  Conference on Artificial Intelligence, IJCAI}}, Vol.~\bibinfo{volume}{5}.
  \bibinfo{pages}{4392--4399}.
\newblock


\bibitem[\protect\citeauthoryear{Dandl, Molnar, Binder, and Bischl}{Dandl
  et~al\mbox{.}}{2020}]%
        {Dandl2020}
\bibfield{author}{\bibinfo{person}{Susanne Dandl}, \bibinfo{person}{Christoph
  Molnar}, \bibinfo{person}{Martin Binder}, {and} \bibinfo{person}{Bernd
  Bischl}.} \bibinfo{year}{2020}\natexlab{}.
\newblock \showarticletitle{{Multi-objective counterfactual explanations}}. In
  \bibinfo{booktitle}{\emph{Proceedings of the 16th International Conference on
  Parallel Problem Solving from Nature}}, Vol.~\bibinfo{volume}{12269 LNCS}.
  \bibinfo{pages}{448--469}.
\newblock


\bibitem[\protect\citeauthoryear{European~Commission}{European~Commission}{2019}]%
        {EuropeanCommission2019}
\bibfield{author}{\bibinfo{person}{High-Level Expert Group on
  Artificial~Intelligence European~Commission}.}
  \bibinfo{year}{2019}\natexlab{}.
\newblock \bibinfo{booktitle}{\emph{Ethics {{Guidelines}} for {{Trustworthy
  AI}}}}.
\newblock \bibinfo{type}{{T}echnical {R}eport}.
\newblock


\bibitem[\protect\citeauthoryear{Fern and Pope}{Fern and Pope}{2021}]%
        {Fern2021}
\bibfield{author}{\bibinfo{person}{Xiaoli Fern} {and} \bibinfo{person}{Quintin
  Pope}.} \bibinfo{year}{2021}\natexlab{}.
\newblock \showarticletitle{Text {{Counterfactuals}} via {{Latent
  Optimization}} and {{Shapley-Guided Search}}}. In
  \bibinfo{booktitle}{\emph{Proceedings of the 2021 {{Conference}} on
  {{Empirical Methods}} in {{Natural Language Processing}}}}.
  \bibinfo{pages}{5578--5593}.
\newblock


\bibitem[\protect\citeauthoryear{Fern{\'{a}}ndez-Lor{\'{i}}a, Provost, and
  Han}{Fern{\'{a}}ndez-Lor{\'{i}}a et~al\mbox{.}}{2021}]%
        {Fernandez-Loria2021}
\bibfield{author}{\bibinfo{person}{Carlos Fern{\'{a}}ndez-Lor{\'{i}}a},
  \bibinfo{person}{Foster Provost}, {and} \bibinfo{person}{Xintian Han}.}
  \bibinfo{year}{2021}\natexlab{}.
\newblock \bibinfo{title}{{Explaining Data-Driven Decisions Explaining
  Data-Driven Decisions made by AI Systems: The Counterfactual Approach}}.
\newblock
\newblock
\showeprint[arxiv]{2001.07417}


\bibitem[\protect\citeauthoryear{{FICO Community}}{{FICO Community}}{2019}]%
        {FICO}
\bibfield{author}{\bibinfo{person}{{FICO Community}}.}
  \bibinfo{year}{2019}\natexlab{}.
\newblock \bibinfo{title}{{Explainable Machine Learning Challenge}}.
\newblock
\newblock
\urldef\tempurl%
\url{https://community.fico.com/s/explainable-machine-learning-challenge}
\showURL{%
\tempurl}


\bibitem[\protect\citeauthoryear{Frye, {de Mijolla}, Begley, Cowton, Stanley,
  and Feige}{Frye et~al\mbox{.}}{2021}]%
        {Frye2021}
\bibfield{author}{\bibinfo{person}{Christopher Frye}, \bibinfo{person}{Damien
  {de Mijolla}}, \bibinfo{person}{Tom Begley}, \bibinfo{person}{Laurence
  Cowton}, \bibinfo{person}{Megan Stanley}, {and} \bibinfo{person}{Ilya
  Feige}.} \bibinfo{year}{2021}\natexlab{}.
\newblock \showarticletitle{Shapley {{Explainability}} on the Data Manifold}.
  In \bibinfo{booktitle}{\emph{Proceedings of the 9th {{International
  Conference}} on {{Learning Representations}} ({{ICLR}})}}.
  \bibinfo{pages}{14}.
\newblock


\bibitem[\protect\citeauthoryear{Guidotti, Monreale, Ruggieri, Pedreschi,
  Turini, and Giannotti}{Guidotti et~al\mbox{.}}{2018}]%
        {Guidotti2018LORE}
\bibfield{author}{\bibinfo{person}{Riccardo Guidotti}, \bibinfo{person}{Anna
  Monreale}, \bibinfo{person}{Salvatore Ruggieri}, \bibinfo{person}{Dino
  Pedreschi}, \bibinfo{person}{Franco Turini}, {and} \bibinfo{person}{Fosca
  Giannotti}.} \bibinfo{year}{2018}\natexlab{}.
\newblock \bibinfo{title}{Local {{Rule}}-{{Based Explanations}} of {{Black Box
  Decision Systems}}}.
\newblock
\newblock
\showeprint[arxiv]{1805.10820}


\bibitem[\protect\citeauthoryear{Guidotti, Monreale, Ruggieri, Turini,
  Giannotti, and Pedreschi}{Guidotti et~al\mbox{.}}{2019}]%
        {Guidotti2019a}
\bibfield{author}{\bibinfo{person}{Riccardo Guidotti}, \bibinfo{person}{Anna
  Monreale}, \bibinfo{person}{Salvatore Ruggieri}, \bibinfo{person}{Franco
  Turini}, \bibinfo{person}{Fosca Giannotti}, {and} \bibinfo{person}{Dino
  Pedreschi}.} \bibinfo{year}{2019}\natexlab{}.
\newblock \showarticletitle{{A Survey of Methods for Explaining Black Box
  Models}}.
\newblock \bibinfo{journal}{\emph{Comput. Surveys}} \bibinfo{volume}{51},
  \bibinfo{number}{5} (\bibinfo{year}{2019}), \bibinfo{pages}{1--42}.
\newblock


\bibitem[\protect\citeauthoryear{Hashemi and Fathi}{Hashemi and Fathi}{2020}]%
        {Hashemi2020}
\bibfield{author}{\bibinfo{person}{Masoud Hashemi} {and} \bibinfo{person}{Ali
  Fathi}.} \bibinfo{year}{2020}\natexlab{}.
\newblock \bibinfo{title}{PermuteAttack: Counterfactual Explanation of Machine
  Learning Credit Scorecards}.
\newblock
\newblock
\showeprint[arxiv]{2008.10138}


\bibitem[\protect\citeauthoryear{Issues}{Issues}{2018}]%
        {SHAPGithub2}
\bibfield{author}{\bibinfo{person}{GitHub Issues}.}
  \bibinfo{year}{2018}\natexlab{}.
\newblock \bibinfo{title}{Interpretation of {{Kernel SHAP}} and Its
  Hyperparameters - {{Issue}} \#23 https://github.com/slundberg/shap}.
\newblock
\newblock


\bibitem[\protect\citeauthoryear{Issues}{Issues}{2019a}]%
        {SHAPGithub4}
\bibfield{author}{\bibinfo{person}{GitHub Issues}.}
  \bibinfo{year}{2019}\natexlab{a}.
\newblock \bibinfo{title}{Choosing the Background Set {$\cdot$} {{Issue}} \#391
  {$\cdot$} https://github.com/slundberg/shap}.
\newblock
\newblock


\bibitem[\protect\citeauthoryear{Issues}{Issues}{2019b}]%
        {SHAPGithub1}
\bibfield{author}{\bibinfo{person}{GitHub Issues}.}
  \bibinfo{year}{2019}\natexlab{b}.
\newblock \bibinfo{title}{Interpretation of {{SHAP Values}} Away from the Mean
  {$\cdot$} {{Issue}} \#435 {$\cdot$} https://github.com/slundberg/shap}.
\newblock
\newblock


\bibitem[\protect\citeauthoryear{Issues}{Issues}{2019c}]%
        {SHAPGithub3}
\bibfield{author}{\bibinfo{person}{GitHub Issues}.}
  \bibinfo{year}{2019}\natexlab{c}.
\newblock \bibinfo{title}{{{ZestFinance}} Writeup on {{SHAP}} and Why It
  Shouldn't Be Used on Its Own {$\cdot$} {{Issue}} \#624 {$\cdot$}
  https://github.com/slundberg/shap}.
\newblock
\newblock


\bibitem[\protect\citeauthoryear{Janzing, Minorics, and Bloebaum}{Janzing
  et~al\mbox{.}}{2020}]%
        {Janzing2020}
\bibfield{author}{\bibinfo{person}{Dominik Janzing}, \bibinfo{person}{Lenon
  Minorics}, {and} \bibinfo{person}{Patrick Bloebaum}.}
  \bibinfo{year}{2020}\natexlab{}.
\newblock \showarticletitle{Feature Relevance Quantification in Explainable
  {{AI}}: {{A}} Causal Problem}. In \bibinfo{booktitle}{\emph{Proceedings of
  the 23rd {{International Conference}} on {{Artificial Intelligence}} and
  {{Statistics}}}}. \bibinfo{pages}{2907--2916}.
\newblock


\bibitem[\protect\citeauthoryear{{Kaggle}}{{Kaggle}}{2019}]%
        {LendingClub}
\bibfield{author}{\bibinfo{person}{{Kaggle}}.} \bibinfo{year}{2019}\natexlab{}.
\newblock \bibinfo{title}{{Lending Club Loan Data}}.
\newblock
\newblock
\urldef\tempurl%
\url{https://www.kaggle.com/wordsforthewise/lending-club}
\showURL{%
\tempurl}


\bibitem[\protect\citeauthoryear{Kanamori, Takagi, Kobayashi, and
  Arimura}{Kanamori et~al\mbox{.}}{2020}]%
        {Kanamori2020}
\bibfield{author}{\bibinfo{person}{Kentaro Kanamori}, \bibinfo{person}{Takuya
  Takagi}, \bibinfo{person}{Ken Kobayashi}, {and} \bibinfo{person}{Hiroki
  Arimura}.} \bibinfo{year}{2020}\natexlab{}.
\newblock \showarticletitle{{{DACE}}: Distribution-{{Aware Counterfactual
  Explanation}} by {{Mixed}}-{{Integer Linear Optimization}}}. In
  \bibinfo{booktitle}{\emph{Proceedings of the 29th International Joint
  Conference on {{Artificial Intelligence}}, IJCAI}}.
  \bibinfo{pages}{2855--2862}.
\newblock


\bibitem[\protect\citeauthoryear{Karimi, Barthe, Balle, and Valera}{Karimi
  et~al\mbox{.}}{2020a}]%
        {Karimi2020}
\bibfield{author}{\bibinfo{person}{Amir-Hossein Karimi},
  \bibinfo{person}{Gilles Barthe}, \bibinfo{person}{Borja Balle}, {and}
  \bibinfo{person}{Isabel Valera}.} \bibinfo{year}{2020}\natexlab{a}.
\newblock \showarticletitle{{Model-Agnostic Counterfactual Explanations for
  Consequential Decisions}}. In \bibinfo{booktitle}{\emph{Proceedings of the
  24th International Conference on Artificial Intelligence and Statistics,
  AISTATS}}. \bibinfo{pages}{895--905}.
\newblock


\bibitem[\protect\citeauthoryear{Karimi, {von K{\"u}gelgen}, Sch{\"o}lkopf, and
  Valera}{Karimi et~al\mbox{.}}{2020b}]%
        {Karimi2020a}
\bibfield{author}{\bibinfo{person}{Amir-Hossein Karimi},
  \bibinfo{person}{Julius {von K{\"u}gelgen}}, \bibinfo{person}{Bernhard
  Sch{\"o}lkopf}, {and} \bibinfo{person}{Isabel Valera}.}
  \bibinfo{year}{2020}\natexlab{b}.
\newblock \showarticletitle{Algorithmic Recourse under Imperfect Causal
  Knowledge: A Probabilistic Approach}. In
  \bibinfo{booktitle}{\emph{{{Proceedings of the 34th Conference on Neural
  Information Processing Systems, NeurIPS}}}}.
\newblock


\bibitem[\protect\citeauthoryear{Karimi, Z{\"{u}}rich, Sch{\"{o}}lkopf, and
  Valera}{Karimi et~al\mbox{.}}{2021}]%
        {Karimi2021d}
\bibfield{author}{\bibinfo{person}{Amir-Hossein Karimi}, \bibinfo{person}{Eth
  Z{\"{u}}rich}, \bibinfo{person}{Switzerland~Bernhard Sch{\"{o}}lkopf}, {and}
  \bibinfo{person}{Isabel Valera}.} \bibinfo{year}{2021}\natexlab{}.
\newblock \showarticletitle{{Algorithmic Recourse: from Counterfactual
  Explanations to Interventions}}. In \bibinfo{booktitle}{\emph{Proceedings of
  the 2021 ACM Conference on Fairness, Accountability, and Transparency,
  FAccT}}. \bibinfo{pages}{353–--362}.
\newblock


\bibitem[\protect\citeauthoryear{Keane, Kenny, Delaney, and Smyth}{Keane
  et~al\mbox{.}}{2021}]%
        {Keane2021}
\bibfield{author}{\bibinfo{person}{Mark~T Keane}, \bibinfo{person}{Eoin~M
  Kenny}, \bibinfo{person}{Eoin Delaney}, {and} \bibinfo{person}{Barry Smyth}.}
  \bibinfo{year}{2021}\natexlab{}.
\newblock \showarticletitle{{If Only We Had Better Counterfactual Explanations:
  Five Key Deficits to Rectify in the Evaluation of Counterfactual XAI
  Techniques}}. In \bibinfo{booktitle}{\emph{Proceeding of the 30th
  International Joint Conference on Artificial Intelligence, IJCAI}}.
  \bibinfo{pages}{4466--4474}.
\newblock


\bibitem[\protect\citeauthoryear{Kumar, Venkatasubramanian, Scheidegger, and
  Friedler}{Kumar et~al\mbox{.}}{2020}]%
        {Kumar2020}
\bibfield{author}{\bibinfo{person}{I~Elizabeth Kumar}, \bibinfo{person}{Suresh
  Venkatasubramanian}, \bibinfo{person}{Carlos Scheidegger}, {and}
  \bibinfo{person}{Sorelle~A Friedler}.} \bibinfo{year}{2020}\natexlab{}.
\newblock \showarticletitle{{Problems with Shapley-value-based explanations as
  feature importance measures}}. In \bibinfo{booktitle}{\emph{Proceedings of
  the 37th International Conference on Machine Learning , ICML}}.
  \bibinfo{pages}{5491--5500}.
\newblock


\bibitem[\protect\citeauthoryear{Lakkaraju, Kamar, Caruana, and
  Leskovec}{Lakkaraju et~al\mbox{.}}{2019}]%
        {Lakkaraju2019}
\bibfield{author}{\bibinfo{person}{Himabindu Lakkaraju}, \bibinfo{person}{Ece
  Kamar}, \bibinfo{person}{Rich Caruana}, {and} \bibinfo{person}{Jure
  Leskovec}.} \bibinfo{year}{2019}\natexlab{}.
\newblock \showarticletitle{Faithful and {{Customizable Explanations}} of
  {{Black Box Models}}}. In \bibinfo{booktitle}{\emph{Proceedings of the 2019
  {{AAAI}}/{{ACM Conference}} on {{AI}}, {{Ethics}}, and {{Society}}, FAccT}}.
  \bibinfo{pages}{131--138}.
\newblock


\bibitem[\protect\citeauthoryear{Laugel, Lesot, Marsala, Renard, and
  Detyniecki}{Laugel et~al\mbox{.}}{2019}]%
        {Laugel2019}
\bibfield{author}{\bibinfo{person}{Thibault Laugel},
  \bibinfo{person}{Marie-Jeanne Lesot}, \bibinfo{person}{Christophe Marsala},
  \bibinfo{person}{Xavier Renard}, {and} \bibinfo{person}{Marcin Detyniecki}.}
  \bibinfo{year}{2019}\natexlab{}.
\newblock \showarticletitle{The Dangers of Post-Hoc Interpretability:
  Unjustified Counterfactual Explanations}. In
  \bibinfo{booktitle}{\emph{Proceedings of the 28th International Joint
  Conference on Artificial Intelligence, IJCAI}}. \bibinfo{pages}{2801--2807}.
\newblock


\bibitem[\protect\citeauthoryear{Lundberg}{Lundberg}{2017}]%
        {Lundberg}
\bibfield{author}{\bibinfo{person}{Scott Lundberg}.}
  \bibinfo{year}{2017}\natexlab{}.
\newblock \bibinfo{title}{Supplementary {{Material}} to a {{Unified Approach}}
  to {{Interpreting Model Predictions}}: {{The}} Monotonicity Axiom Implies the
  Symmetry Axiom for {{Shapley}} Values}.
\newblock
\newblock


\bibitem[\protect\citeauthoryear{Lundberg, Erion, Chen, DeGrave, Prutkin, Nair,
  Katz, Himmelfarb, Bansal, and Lee}{Lundberg et~al\mbox{.}}{2020}]%
        {Lundberg2020Trees}
\bibfield{author}{\bibinfo{person}{Scott~M. Lundberg}, \bibinfo{person}{Gabriel
  Erion}, \bibinfo{person}{Hugh Chen}, \bibinfo{person}{Alex DeGrave},
  \bibinfo{person}{Jordan~M. Prutkin}, \bibinfo{person}{Bala Nair},
  \bibinfo{person}{Ronit Katz}, \bibinfo{person}{Jonathan Himmelfarb},
  \bibinfo{person}{Nisha Bansal}, {and} \bibinfo{person}{Su-In Lee}.}
  \bibinfo{year}{2020}\natexlab{}.
\newblock \showarticletitle{{From local explanations to global understanding
  with explainable AI for trees}}.
\newblock \bibinfo{journal}{\emph{Nature Machine Intelligence}}
  \bibinfo{volume}{2}, \bibinfo{number}{1} (\bibinfo{date}{1}
  \bibinfo{year}{2020}), \bibinfo{pages}{56--67}.
\newblock


\bibitem[\protect\citeauthoryear{Lundberg and Lee}{Lundberg and Lee}{2017}]%
        {Lundberg17}
\bibfield{author}{\bibinfo{person}{Scott~M Lundberg} {and}
  \bibinfo{person}{Su-In Lee}.} \bibinfo{year}{2017}\natexlab{}.
\newblock \showarticletitle{{A Unified Approach to Interpreting Model
  Predictions}}. In \bibinfo{booktitle}{\emph{Advances in Neural Information
  Processing Systems, NeurIPS}}. \bibinfo{pages}{4768--–4777}.
\newblock


\bibitem[\protect\citeauthoryear{Merrick and Taly}{Merrick and Taly}{2020}]%
        {Merrick2020}
\bibfield{author}{\bibinfo{person}{Luke Merrick} {and} \bibinfo{person}{Ankur
  Taly}.} \bibinfo{year}{2020}\natexlab{}.
\newblock \showarticletitle{{The Explanation Game: Explaining Machine Learning
  Models Using Shapley Values}}. In \bibinfo{booktitle}{\emph{International
  Cross-Domain Conference for Machine Learning and Knowledge Extraction,
  CD-MAKE}}. \bibinfo{pages}{17--38}.
\newblock


\bibitem[\protect\citeauthoryear{Merrill, Ward, Kamkar, Budzik, and
  Merrill}{Merrill et~al\mbox{.}}{2019}]%
        {Merrill2019}
\bibfield{author}{\bibinfo{person}{John W~L Merrill}, \bibinfo{person}{Geoff~M
  Ward}, \bibinfo{person}{Sean~J Kamkar}, \bibinfo{person}{Jay Budzik}, {and}
  \bibinfo{person}{Douglas~C Merrill}.} \bibinfo{year}{2019}\natexlab{}.
\newblock \showarticletitle{{Generalized Integrated Gradients: A practical
  method for explaining diverse ensembles}}.
\newblock \bibinfo{journal}{\emph{Journal of Machine Learning Research - Under
  Review}} (\bibinfo{year}{2019}).
\newblock
\showeprint[arxiv]{1909.01869}


\bibitem[\protect\citeauthoryear{Miller}{Miller}{2019}]%
        {Miller2019}
\bibfield{author}{\bibinfo{person}{Tim Miller}.}
  \bibinfo{year}{2019}\natexlab{}.
\newblock \showarticletitle{{Explanation in Artificial Intelligence: Insights
  from the Social Sciences}}.
\newblock \bibinfo{journal}{\emph{Artificial Intelligence}}
  \bibinfo{volume}{267} (\bibinfo{date}{6} \bibinfo{year}{2019}),
  \bibinfo{pages}{1--38}.
\newblock


\bibitem[\protect\citeauthoryear{Mothilal, Mahajan, Tan, and Sharma}{Mothilal
  et~al\mbox{.}}{2021}]%
        {Mothilal2021}
\bibfield{author}{\bibinfo{person}{R.~K. Mothilal}, \bibinfo{person}{Divyat
  Mahajan}, \bibinfo{person}{Chenhao Tan}, {and} \bibinfo{person}{Amit
  Sharma}.} \bibinfo{year}{2021}\natexlab{}.
\newblock \showarticletitle{Towards {{Unifying Feature Attribution}} and
  {{Counterfactual Explanations}}: Different {{Means}} to the {{Same End}}}. In
  \bibinfo{booktitle}{\emph{Proceedings of the 2021 {{AAAI}}/{{ACM Conference}}
  on {{AI}}, {{Ethics}}, and Society, AIES}}. \bibinfo{pages}{652--663}.
\newblock


\bibitem[\protect\citeauthoryear{Mothilal, Sharma, and Tan}{Mothilal
  et~al\mbox{.}}{2020}]%
        {Mothilal2020a}
\bibfield{author}{\bibinfo{person}{Ramaravind~K. Mothilal},
  \bibinfo{person}{Amit Sharma}, {and} \bibinfo{person}{Chenhao Tan}.}
  \bibinfo{year}{2020}\natexlab{}.
\newblock \showarticletitle{Explaining Machine Learning Classifiers through
  Diverse Counterfactual Explanations}. In
  \bibinfo{booktitle}{\emph{Proceedings of the 2020 {{Conference}} on
  {{Fairness}}, {{Accountability}}, and {{Transparency}}, FAccT}}.
  \bibinfo{pages}{607--617}.
\newblock


\bibitem[\protect\citeauthoryear{Pawelczyk, Bielawski, van~den Heuvel, Richter,
  and Kasneci}{Pawelczyk et~al\mbox{.}}{2021}]%
        {Pawelczyk2021}
\bibfield{author}{\bibinfo{person}{Martin Pawelczyk}, \bibinfo{person}{Sascha
  Bielawski}, \bibinfo{person}{Johannes van~den Heuvel},
  \bibinfo{person}{Tobias Richter}, {and} \bibinfo{person}{Gjergji Kasneci}.}
  \bibinfo{year}{2021}\natexlab{}.
\newblock \showarticletitle{{{CARLA}}: {{A Python Library}} to {{Benchmark
  Algorithmic Recourse}} and {{Counterfactual Explanation Algorithms}}}. In
  \bibinfo{booktitle}{\emph{Benchmark \& Data Sets Track at the 36th Conference
  on Neural Information Processing Systems, NeurIPS}}.
\newblock


\bibitem[\protect\citeauthoryear{Pawelczyk, Broelemann, and Kasneci}{Pawelczyk
  et~al\mbox{.}}{2020}]%
        {Pawelczyk2020}
\bibfield{author}{\bibinfo{person}{Martin Pawelczyk}, \bibinfo{person}{Klaus
  Broelemann}, {and} \bibinfo{person}{Gjergji Kasneci}.}
  \bibinfo{year}{2020}\natexlab{}.
\newblock \showarticletitle{{Learning Model-Agnostic Counterfactual
  Explanations for Tabular Data}}. In \bibinfo{booktitle}{\emph{Proceedings of
  The Web Conference 2020, WWW}}. \bibinfo{pages}{3126–--3132}.
\newblock


\bibitem[\protect\citeauthoryear{Plumb, Molitor, and Talwalkar}{Plumb
  et~al\mbox{.}}{2018}]%
        {Plumb2018}
\bibfield{author}{\bibinfo{person}{Gregory Plumb}, \bibinfo{person}{Denali
  Molitor}, {and} \bibinfo{person}{Ameet~S Talwalkar}.}
  \bibinfo{year}{2018}\natexlab{}.
\newblock \showarticletitle{Model {{Agnostic Supervised Local Explanations}}}.
  In \bibinfo{booktitle}{\emph{Advances in {{Neural Information Processing
  Systems}}, NeurIPS}}. \bibinfo{pages}{2520–--2529}.
\newblock


\bibitem[\protect\citeauthoryear{Poyiadzi, Sokol, Santos-Rodriguez, De~Bie, and
  Flach}{Poyiadzi et~al\mbox{.}}{2020}]%
        {Poyiadzi2019}
\bibfield{author}{\bibinfo{person}{Rafael Poyiadzi}, \bibinfo{person}{Kacper
  Sokol}, \bibinfo{person}{Raul Santos-Rodriguez}, \bibinfo{person}{Tijl
  De~Bie}, {and} \bibinfo{person}{Peter Flach}.}
  \bibinfo{year}{2020}\natexlab{}.
\newblock \showarticletitle{{FACE: Feasible and Actionable Counterfactual
  Explanations}}. In \bibinfo{booktitle}{\emph{Proceedings of the AAAI/ACM
  Conference on AI, Ethics, and Society, AIES}}. \bibinfo{pages}{344--350}.
\newblock


\bibitem[\protect\citeauthoryear{Ramon, Martens, Provost, and Evgeniou}{Ramon
  et~al\mbox{.}}{2020}]%
        {Ramon2020}
\bibfield{author}{\bibinfo{person}{Yanou Ramon}, \bibinfo{person}{David
  Martens}, \bibinfo{person}{Foster Provost}, {and} \bibinfo{person}{Theodoros
  Evgeniou}.} \bibinfo{year}{2020}\natexlab{}.
\newblock \showarticletitle{{A comparison of instance-level counterfactual
  explanation algorithms for behavioral and textual data: SEDC, LIME-C and
  SHAP-C}}.
\newblock \bibinfo{journal}{\emph{Advances in Data Analysis and
  Classification}}  \bibinfo{volume}{14} (\bibinfo{year}{2020}),
  \bibinfo{pages}{801--819}.
\newblock


\bibitem[\protect\citeauthoryear{Rathi}{Rathi}{2019}]%
        {Rathi2019}
\bibfield{author}{\bibinfo{person}{Shubham Rathi}.}
  \bibinfo{year}{2019}\natexlab{}.
\newblock \showarticletitle{{Generating Counterfactual and Contrastive
  Explanations using SHAP}}. In \bibinfo{booktitle}{\emph{2nd Workshop on
  Humanizing AI (HAI) at IJCAI '19}}.
\newblock
\showeprint[arxiv]{1906.09293}


\bibitem[\protect\citeauthoryear{Rawal and Lakkaraju}{Rawal and
  Lakkaraju}{2020}]%
        {Rawal2020}
\bibfield{author}{\bibinfo{person}{Kaivalya Rawal} {and}
  \bibinfo{person}{Himabindu Lakkaraju}.} \bibinfo{year}{2020}\natexlab{}.
\newblock \showarticletitle{{Beyond Individualized Recourse: Interpretable and
  Interactive Summaries of Actionable Recourses}}. In
  \bibinfo{booktitle}{\emph{Advances in Neural Information Processing Systems,
  NeurIPS}}. \bibinfo{pages}{12187--12198}.
\newblock


\bibitem[\protect\citeauthoryear{Ribeiro, Singh, and Guestrin}{Ribeiro
  et~al\mbox{.}}{2016}]%
        {Ribeiro2016}
\bibfield{author}{\bibinfo{person}{Marco~Tulio Ribeiro},
  \bibinfo{person}{Sameer Singh}, {and} \bibinfo{person}{Carlos Guestrin}.}
  \bibinfo{year}{2016}\natexlab{}.
\newblock \showarticletitle{{"Why Should I Trust You?"}}. In
  \bibinfo{booktitle}{\emph{Proceedings of the 22nd ACM SIGKDD International
  Conference on Knowledge Discovery and Data Mining, KDD}}.
  \bibinfo{pages}{1135--1144}.
\newblock


\bibitem[\protect\citeauthoryear{Roth}{Roth}{1988}]%
        {Shapley1988}
\bibfield{author}{\bibinfo{person}{Alvin~E. Roth}.}
  \bibinfo{year}{1988}\natexlab{}.
\newblock \bibinfo{booktitle}{\emph{The {{Shapley}} Value: Essays in Honor of
  {{Lloyd S}}. {{Shapley}}}}.
\newblock
\showLCCN{HB144 .S533 1988}


\bibitem[\protect\citeauthoryear{Russell}{Russell}{2019}]%
        {Russell2019}
\bibfield{author}{\bibinfo{person}{Chris Russell}.}
  \bibinfo{year}{2019}\natexlab{}.
\newblock \showarticletitle{{Efficient Search for Diverse Coherent
  Explanations}}. In \bibinfo{booktitle}{\emph{Proceedings of the 2019
  Conference on Fairness, Accountability, and Transparency, FAccT}}.
  \bibinfo{pages}{20--28}.
\newblock


\bibitem[\protect\citeauthoryear{Shapley}{Shapley}{1951}]%
        {Shapley1951}
\bibfield{author}{\bibinfo{person}{Lloyd~Stowell Shapley}.}
  \bibinfo{year}{1951}\natexlab{}.
\newblock \bibinfo{title}{{Notes on the n-Person Game-II: The Value of an
  n-Person Game}}.
\newblock
\newblock


\bibitem[\protect\citeauthoryear{Sharma, Henderson, and Ghosh}{Sharma
  et~al\mbox{.}}{2020}]%
        {Sharma2020}
\bibfield{author}{\bibinfo{person}{Shubham Sharma}, \bibinfo{person}{Jette
  Henderson}, {and} \bibinfo{person}{Joydeep Ghosh}.}
  \bibinfo{year}{2020}\natexlab{}.
\newblock \showarticletitle{{CERTIFAI: A Common Framework to Provide
  Explanations and Analyse the Fairness and Robustness of Black-box Models}}.
  In \bibinfo{booktitle}{\emph{Proceedings of the AAAI/ACM Conference on AI,
  Ethics, and Society, AIES}}. \bibinfo{pages}{166--172}.
\newblock


\bibitem[\protect\citeauthoryear{{Shwartz-Ziv} and Armon}{{Shwartz-Ziv} and
  Armon}{2021}]%
        {Shwartz-Ziv2021}
\bibfield{author}{\bibinfo{person}{Ravid {Shwartz-Ziv}} {and}
  \bibinfo{person}{Amitai Armon}.} \bibinfo{year}{2021}\natexlab{}.
\newblock \bibinfo{title}{Tabular {{Data}}: {{Deep Learning}} Is {{Not All You
  Need}}}.
\newblock
\newblock
\showeprint[arxiv]{2106.03253}


\bibitem[\protect\citeauthoryear{Smyth and Keane}{Smyth and Keane}{2021}]%
        {Smyth}
\bibfield{author}{\bibinfo{person}{Barry Smyth} {and} \bibinfo{person}{Mark~T.
  Keane}.} \bibinfo{year}{2021}\natexlab{}.
\newblock \bibinfo{title}{A {{Few Good Counterfactuals}}: Generating
  {{Interpretable}}, {{Plausible}} and {{Diverse Counterfactual
  Explanations}}}.
\newblock
\newblock
\showeprint[arxiv]{2101.09056}


\bibitem[\protect\citeauthoryear{Spooner, Dervovic, Long, Shepard, Chen, and
  Magazzeni}{Spooner et~al\mbox{.}}{2021}]%
        {Spooner2021}
\bibfield{author}{\bibinfo{person}{Thomas Spooner}, \bibinfo{person}{Danial
  Dervovic}, \bibinfo{person}{Jason Long}, \bibinfo{person}{Jon Shepard},
  \bibinfo{person}{Jiahao Chen}, {and} \bibinfo{person}{Daniele Magazzeni}.}
  \bibinfo{year}{2021}\natexlab{}.
\newblock \showarticletitle{Counterfactual {{Explanations}} for {{Arbitrary
  Regression Models}}}. In \bibinfo{booktitle}{\emph{{{ICML}}'21 {{Workshop}}
  on {{Algorithmic Recourse}}}}.
\newblock
\showeprint[arxiv]{2106.15212}


\bibitem[\protect\citeauthoryear{Stepin, Alonso, Catala, and
  Pereira-Farina}{Stepin et~al\mbox{.}}{2021}]%
        {Stepin2021}
\bibfield{author}{\bibinfo{person}{Ilia Stepin}, \bibinfo{person}{Jose~M.
  Alonso}, \bibinfo{person}{Alejandro Catala}, {and} \bibinfo{person}{Martin
  Pereira-Farina}.} \bibinfo{year}{2021}\natexlab{}.
\newblock \showarticletitle{{A Survey of Contrastive and Counterfactual
  Explanation Generation Methods for Explainable Artificial Intelligence}}.
\newblock \bibinfo{journal}{\emph{IEEE Access}}  \bibinfo{volume}{9}
  (\bibinfo{year}{2021}), \bibinfo{pages}{11974--12001}.
\newblock


\bibitem[\protect\citeauthoryear{Strumbelj and Kononenko}{Strumbelj and
  Kononenko}{2010}]%
        {Strumbelj2010}
\bibfield{author}{\bibinfo{person}{Erik Strumbelj} {and} \bibinfo{person}{Igor
  Kononenko}.} \bibinfo{year}{2010}\natexlab{}.
\newblock \showarticletitle{{An Efficient Explanation of Individual
  Classifications using Game Theory}}.
\newblock \bibinfo{journal}{\emph{Journal of Machine Learning Research}}
  \bibinfo{volume}{11} (\bibinfo{year}{2010}), \bibinfo{pages}{1--18}.
\newblock


\bibitem[\protect\citeauthoryear{Sudjianto and Zoldi}{Sudjianto and
  Zoldi}{2021}]%
        {Sudjianto2021Podcast}
\bibfield{author}{\bibinfo{person}{Agus Sudjianto} {and} \bibinfo{person}{Scott
  Zoldi}.} \bibinfo{year}{2021}\natexlab{}.
\newblock \bibinfo{title}{The {Case} for {Interpretable} {Models} in {Credit}
  {Underwriting}}.
\newblock
\newblock
\urldef\tempurl%
\url{https://soundcloud.com/finreglab/agus-sudjiantoscott-zoldi-the-case-for-interpretable-models-in-credit-underwriting}
\showURL{%
\tempurl}


\bibitem[\protect\citeauthoryear{Sundararajan and Najmi}{Sundararajan and
  Najmi}{2020}]%
        {Sundararajan2020}
\bibfield{author}{\bibinfo{person}{Mukund Sundararajan} {and}
  \bibinfo{person}{Amir Najmi}.} \bibinfo{year}{2020}\natexlab{}.
\newblock \showarticletitle{The {{Many Shapley Values}} for {{Model
  Explanation}}}. In \bibinfo{booktitle}{\emph{Proceedings of the 37th
  {{International Conference}} on {{Machine Learning}}}}.
  \bibinfo{pages}{9269--9278}.
\newblock


\bibitem[\protect\citeauthoryear{{U.S. Congress}}{{U.S. Congress}}{2018}]%
        {EqualOpportunityAct}
\bibfield{author}{\bibinfo{person}{{U.S. Congress}}.}
  \bibinfo{year}{2018}\natexlab{}.
\newblock \bibinfo{title}{{12 CFR Part 1002 - Equal Credit Opportunity Act
  (Regulation B)}}.
\newblock
\newblock
\urldef\tempurl%
\url{https://www.consumerfinance.gov/rules-policy/regulations/1002/9/}
\showURL{%
\tempurl}


\bibitem[\protect\citeauthoryear{Ustun, Spangher, and Liu}{Ustun
  et~al\mbox{.}}{2019}]%
        {Ustun}
\bibfield{author}{\bibinfo{person}{Berk Ustun}, \bibinfo{person}{Alexander
  Spangher}, {and} \bibinfo{person}{Yang Liu}.}
  \bibinfo{year}{2019}\natexlab{}.
\newblock \showarticletitle{{Actionable Recourse in Linear Classification}}. In
  \bibinfo{booktitle}{\emph{Proceedings of the Conference on Fairness,
  Accountability, and Transparency, FAccT}}. \bibinfo{pages}{10--19}.
\newblock


\bibitem[\protect\citeauthoryear{Verma, Ai, Dickerson, and Hines}{Verma
  et~al\mbox{.}}{2020}]%
        {Verma}
\bibfield{author}{\bibinfo{person}{Sahil Verma}, \bibinfo{person}{Arthur Ai},
  \bibinfo{person}{John Dickerson}, {and} \bibinfo{person}{Keegan Hines}.}
  \bibinfo{year}{2020}\natexlab{}.
\newblock \bibinfo{title}{{Counterfactual Explanations for Machine Learning: A
  Review}}.
\newblock
\newblock
\showeprint[arxiv]{2010.10596}


\bibitem[\protect\citeauthoryear{{von K{\"u}gelgen}, Karimi, Bhatt, Valera,
  Weller, and Sch{\"o}lkopf}{{von K{\"u}gelgen} et~al\mbox{.}}{2022}]%
        {vonKugelgen2022}
\bibfield{author}{\bibinfo{person}{Julius {von K{\"u}gelgen}},
  \bibinfo{person}{Amir-Hossein Karimi}, \bibinfo{person}{Umang Bhatt},
  \bibinfo{person}{Isabel Valera}, \bibinfo{person}{Adrian Weller}, {and}
  \bibinfo{person}{Bernhard Sch{\"o}lkopf}.} \bibinfo{year}{2022}\natexlab{}.
\newblock \showarticletitle{On the {{Fairness}} of {{Causal Algorithmic
  Recourse}}}. In \bibinfo{booktitle}{\emph{Proceedings of the 36th {{AAAI
  Conference}} on {{Artificial Intelligence}}}}.
\newblock


\bibitem[\protect\citeauthoryear{Wachter, Mittelstadt, and Russell}{Wachter
  et~al\mbox{.}}{2018}]%
        {Wachter2017}
\bibfield{author}{\bibinfo{person}{Sandra Wachter}, \bibinfo{person}{Brent
  Mittelstadt}, {and} \bibinfo{person}{Chris Russell}.}
  \bibinfo{year}{2018}\natexlab{}.
\newblock \showarticletitle{{Counterfactual Explanations Without Opening the
  Black Box: Automated Decisions and the GDPR}}.
\newblock \bibinfo{journal}{\emph{Harvard Journal of Law \& Technology}}
  \bibinfo{volume}{31} (\bibinfo{year}{2018}), \bibinfo{pages}{1--52}.
\newblock


\bibitem[\protect\citeauthoryear{Wang, Wiens, and Lundberg}{Wang
  et~al\mbox{.}}{2021}]%
        {Wang2021}
\bibfield{author}{\bibinfo{person}{Jiaxuan Wang}, \bibinfo{person}{Jenna
  Wiens}, {and} \bibinfo{person}{Scott Lundberg}.}
  \bibinfo{year}{2021}\natexlab{}.
\newblock \showarticletitle{Shapley {{Flow}}: {{A Graph-based Approach}} to
  {{Interpreting Model Predictions}}}. In \bibinfo{booktitle}{\emph{Proocedings
  of the 24th {{International Conference}} on {{Artificial Intelligence}} and
  {{Statistics}}, AISTATS}}.
\newblock


\bibitem[\protect\citeauthoryear{Watson}{Watson}{2022}]%
        {Watson2021}
\bibfield{author}{\bibinfo{person}{David~S. Watson}.}
  \bibinfo{year}{2022}\natexlab{}.
\newblock \showarticletitle{Rational {{Shapley Values}}}. In
  \bibinfo{booktitle}{\emph{Proceedings of the 2022 Conference on Fairness,
  Accountability, and Transparency, FAccT}}.
\newblock


\bibitem[\protect\citeauthoryear{White and Garcez}{White and Garcez}{2019}]%
        {White2019}
\bibfield{author}{\bibinfo{person}{Adam White} {and}
  \bibinfo{person}{Artur~d'Avila Garcez}.} \bibinfo{year}{2019}\natexlab{}.
\newblock \showarticletitle{{Measurable Counterfactual Local Explanations for
  Any Classifier}}. In \bibinfo{booktitle}{\emph{Proceedings of the 24th
  European Conference on Artificial Intelligence, ECAI}}.
  \bibinfo{pages}{2529--2535}.
\newblock


\end{thebibliography}

\pagebreak
$\quad$
\pagebreak
\appendix
\begin{table*}[ht!]
\begin{center}
    \begin{small}
        \begin{tabular}{rcccccccc}
            \toprule
                \multirow{2}{*}{Dataset} &
                \multicolumn{3}{c}{Size} &
                Decision &
                \multicolumn{3}{c}{Model Performance$^{\dagger}$}\\
                & Features\! & \!Train Set\! & \!Test Set & Threshold$^{*}$ & ROC-AUC\! & \!Recall\! & \!Accuracy$^{\ddagger}$\! \\
            \midrule
                \!\textbf{HELOC} (Home Equity Line Of Credit) \cite{FICO}  & 23 & 6,909 & 2,962        & 0.3985 & 79.6\% & 81.6\% & 72.9\% \\
                \!\textbf{LC} (Lending Club Loan Data) \cite{LendingClub}  & 20 & 961,326 & 411,998    & 0.3824 & 69.6\% & 79.8\% & 56.0\% \\ 
                \!\textbf{WINE} (UCI Wine Quality) \cite{Cortez2009}       & 11 & 3,428 & 1,470        & 0.4614 & 83.2\% & 80.7\% & 78.2\% \\
            \bottomrule
        \end{tabular}
    \end{small}
\end{center}
    \protect\caption{\EB{
    Characteristics of the datasets and models used in the experiments.
    ($*$) The decision threshold is reported here in probability space (i.e., after passing the model output through a sigmoid);
    ($\dagger$) performance metrics are computed on the test set;
    ($\ddagger$) we note that the AUC-ROC and recall are better suited metrics for this applications (i.e., a ``bad'' customer being accepted is a more undesirable outcome than a ``good'' customer being rejected).
    }}
    \label{table:datasets}
\end{table*}

\section{Experimental Setup and Reproducibility}
\label{appendix:setup}

\subsection{Datasets and Models}
To run the experiments we used 3 publicly available datasets. Table~\ref{table:datasets} describes in details the datasets.

We split the data using a stratified $70/30$ \emph{random} train/test split for HELOC and WINE. For LC we split the data using a non-random $70/30$ train/test split based on the loan issuance date (available in the original data).

We trained an XGBoost model \cite{XGBoost} for each dataset. In particular, we hyper-trained the parameters using Bayesian optimization via hyperopt~\cite{hyperopt} for $2000$ iterations maximizing the average validation ROC-AUC under a 5-fold cross validation.
To reduce model over-parameterization during the hyper-parameters optimization we penalized high model variance, i.e., for each cross-validation fold, instead of using $AUC_{val}$, we used $AUC_{val} + (AUC_{val} - AUC_{train})$ where $AUC_{train}$ and $AUC_{val}$ are the training and validation ROC-AUC, respectively.

To compute the decision threshold ($t$) we used the ROC-AUC curve: we maximized the sum of the false positive rate (fall-out) and true positive rate (recall).
Table~\ref{table:datasets} shows the decision threshold and the performance of each model.

\subsection{Technical setup}
The experiments were run using a \texttt{c6i.8xlarge} AWS virtual machine with 32 vCPUs (16 cores of 3.5 GHz 3rd generation Intel Xeon Scalable processor) and 64GB of RAM. XGBoost parameter \texttt{nthread} was set to \texttt{15}.

We used a machine running \texttt{Ubuntu 20.04}. We used \texttt{Python 3.6.13}, \texttt{shap 0.39.0}, \texttt{sklearn 0.24.2} and \texttt{xgboost 1.3.3}.

\subsection{Source Code}
The code to reproduce the experiments will be made available at  \url{{https://www.emanuelealbini.com/cfshap-facct2022}}.

\subsubsection{SHAP Explanations}
In order to compute Shapley values we used the TreeSHAP \cite{Lundberg2020Trees} available through the \texttt{TreeExplainer} class in the \texttt{shap} package\footnote{The \texttt{shap} package can be found at \url{https://github.com/slundberg/shap}} (for Python).
We note that we computed the Shapley values on the model output (default setting of \texttt{shap}). We also remark that we used the interventional (a.k.a., non-conditional) version of SHAP (default setting of \texttt{shap}).

\subsubsection{K-Nearest Neighbours}
To compute the $K$-nearest neighbours implementation in \texttt{sklearn.neighbours}. \ED{To make our results indifferent to the size of the dataset we limited the $k$-nearest neighbours to be selected among a random sample of $10,000$ samples from the training set.}

\section{Counterfactual SHAP algorithm: properties and execution time}\label{appendix:performance}\label{appendix:algorithm}

\begin{algorithm*}
    \begin{algorithmic}
        \Procedure{CF-SHAP}{$x$, $f$}
            \State $D_{C} \gets C(\x, f)$ \Comment{The set of counterfactuals $D_{C}$ is computed for $x$ wrt. the model $f$ using technique $C$.}
            \State $\phiv \gets SHAP(\x, f, D_{C})$ \Comment{The Shapley values $\phiv$ for $\x$ wrt. background dataset $D_C$ and model $f$ using SHAP.}
            \State $\tauv \gets Trends(\x, D_{C})$ \Comment{The trends $\tauv$ for $\x$ wrt. background dataset $D_C$ are computed.}
            \State \Return $(\phiv, \tauv)$ \Comment{Shapley values and trends are returned.}
        \EndProcedure
    \end{algorithmic}
    \caption{Counterfactual SHAP algorithm}
    \label{algo:cshap}
\end{algorithm*}

We report in Algorithm~\ref{algo:cshap} the procedure to compute Counterfactual SHAP explanations.

\subsection{Properties of Counterfactual SHAP}

We will now discuss some of the properties of Counterfactual SHAP.

We note that Counterfactual SHAP is an \emph{additive feature attribution method} as defined by \cite{Lundberg17} (or, equivalently, it satisfies \emph{additivity}), i.e., $f(\x) = \phi_0 + \sum_{i\in\F}\phi_i$ where $\phi_0 = E_{\x'\in\D_C(f, \x)}\left[f(\x')\right]$. 
We also note that Counterfactual SHAP values satisfy all the properties (local accuracy, missingness and consistency) satisfied by SHAP values.
This is true because such properties are satisfied by SHAP values \textbf{independently of the background distribution}.
In fact, the definition of Counterfactual SHAP values diverges from the definition of SHAP values only in terms of the background distribution used in the characteristic function.\footnote{\EB{In \cite{Lundberg17} the characteristic function is denoted with $f_x$ while in this paper we use the canonical notation $v$ from the game theory literature.}}

\textbf{Additivity}. As noted in Section~\ref{sec:counterfactual-explanations} one of the objectives of counterfactual generation techniques is to try to minimize the distance between the input $\x$ and the counterfactual point $\x'$ and indirectly the distance of $\x'$ from the decision boundary.
Since Counterfactual SHAP uses a set of counterfactual points as background distribution, we note that the closer (in terms of output) the generated counterfactuals are to the decision boundary, the better the sum of the Shapley values approximates the distance in model output of the query instance from the decision boundary. Indeed, rewriting the additivity property for Counterfactual SHAP we have
$$
\sum_{i\in\F} \phi_i = f(\x) - \E_{\x'\in\D_C(\x,f)}\left[f(\x')\right].
$$

\begin{figure*}
\centering
\includegraphics[width=.95\textwidth]{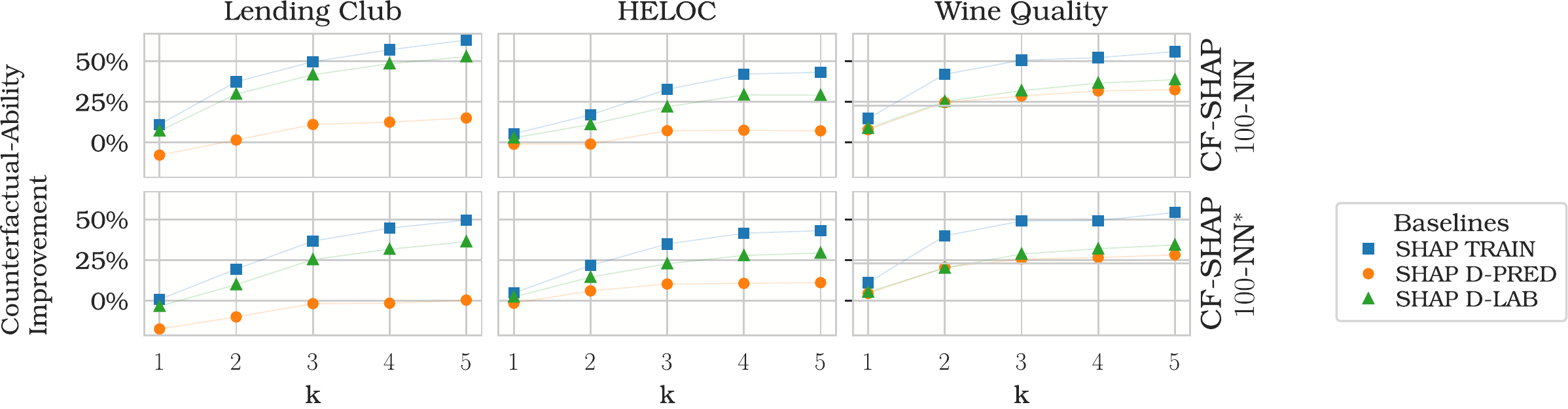}
\caption{\EF{Counterfactual-ability improvement (as defined in Section~\ref{sec:experiments}) of CF-SHAP $100$-NN and CF-SHAP $100$-NN$*$ with respect to the baselines. We note that CF-SHAP $100$-NN and CF-SHAP $100$-NN$*$ have very similar performance in terms of counterfactual-ability. ($*$) indicates the variant of CF-SHAP $K$-NN that generates counterfactuals by projecting the $K$-nearest neighbours onto the decision boundary along the line to the query instance.}}
\label{fig:addtivity_knn}
\end{figure*}

When using Counterfactual SHAP the average model output on the counterfactual distribution, $\E_{\x'\in\D_C(\x,f)} \left[f(\x') \right]$, should approximate the threshold output value $t$, although this does depend on properties of the model and the counterfactual distribution. If the model is continuous and the counterfactuals are selected to be on the decision boundary then we will have $\E_{\x'\in\D_C(\x,f)}\left[f(\x')\right]=t$, but this cannot be guaranteed theoretically for tree-based models. Empirically we see that $\E_{\x'\in\D_C(\x,f)}\left[f(\x')\right]$ approximates $t$ very closely for certain choices of counterfactual distribution. 
For example, by projecting the $K$-nearest neighbours onto the decision boundary along the line to the query instance, we can obtain a counterfactual distribution $D_C(\x,f)$ with $\E_{\x'\in\D_C(\x,f)}\left[f(\x')\right]$ very close to the threshold (see Table~\ref{table:additivity_experiment}). Using this counterfactual distribution instead of the $K$-nearest neighbours themselves results in extremely similar performance to that presented in Section~\ref{sec:experiments}, as shown in Figure~\ref{fig:addtivity_knn}.

\begin{table}[!t]
\begin{tabular}{rlll}
\toprule
 &      HELOC &        LC &        WQ \\
\midrule
CF-SHAP $100$-NN         &   0.524610 &  0.268826 &  0.663118 \\
CF-SHAP $100$-NN $^*$    &   \textbf{0.067102} &  0.080836 &  \textbf{0.076111} \\
SHAP D-LAB                 &   0.330960 &  \textbf{0.037210} &  0.567720 \\
SHAP D-PRED                &   0.940211 &  0.693693 &  0.880631 \\
SHAP TRAIN                 &   0.437677 &  0.102876 &  0.456406 \\
\bottomrule
\end{tabular}
    \protect\caption{\EB{
        Divergence of the average model output of the points in the background dataset from the threshold $t$ for different distributions, i.e., $\left| t - \E_{\x \in \D}\left[ f(\x) \right] \right|$ where $\D$ is the background distribution used to compute the Shapley values. \EF{($*$) indicates the variant of CF-SHAP $K$-NN that generates counterfactuals by projecting the $K$-nearest neighbours onto the decision boundary along the line to the query instance.}
    }}
    \label{table:additivity_experiment}
\end{table}


\textbf{\EC{Linearity}}\footnote{\EC{The linearity property} should not be confused with additivity used earlier in this paper (a.k.a., local accuracy, or efficiency), i.e., $f(\x) = \sum_{i\in \{0\}\cup\F}\phi_i$.}.
For Shapley values, linearity states that given coalitional games $\Gamma$ and $\Gamma'$ with value functions $v$ and $w$ respectively, then the Shapley values $\phi_i(v+w)$ for the summed game $\Gamma+\Gamma'$ with value $v+w$ are given by the sum $\phi_i(v)+\phi_i(w)$ of the Shapley values of $\Gamma$ and $\Gamma'$. In the context of machine learning models, this axiom states that for two models $F_1$ and $F_2$ and a {\bf fixed background dataset} the SHAP values for the summed model $f_\Sigma=f_1+f_2$ are given by the sum of the SHAP values for $f_1$ and $f_2$. This property is essential for rapidly calculating the SHAP values for ensemble models, as for example in the TreeSHAP algorithm~\cite{Lundberg2020Trees}. 

CF-SHAP involves the calculation of SHAP values, and as such the linearity property as described above is inherited automatically for this calculation (allowing the use of efficient algorithms such as TreeSHAP). However, the use of a fixed background dataset is an important caveat. In CF-SHAP it is key that the background dataset is tailored both to the individual query instance and also to the model. Counterfactuals for one model may not be counterfactuals for another model on the same dataset, and it is therefore not the case that the CF-SHAP values for $f_1+f_2$ are given by summing the CF-SHAP values for $f_1$ and $f_2$. 
Thus, when viewing the Counterfactual SHAP algorithm as a whole, we see that the linearity property will typically not hold since the counterfactual distributions will differ for different models.

\begin{figure*}[!ht]
\centering
\includegraphics[height=.35\textwidth]{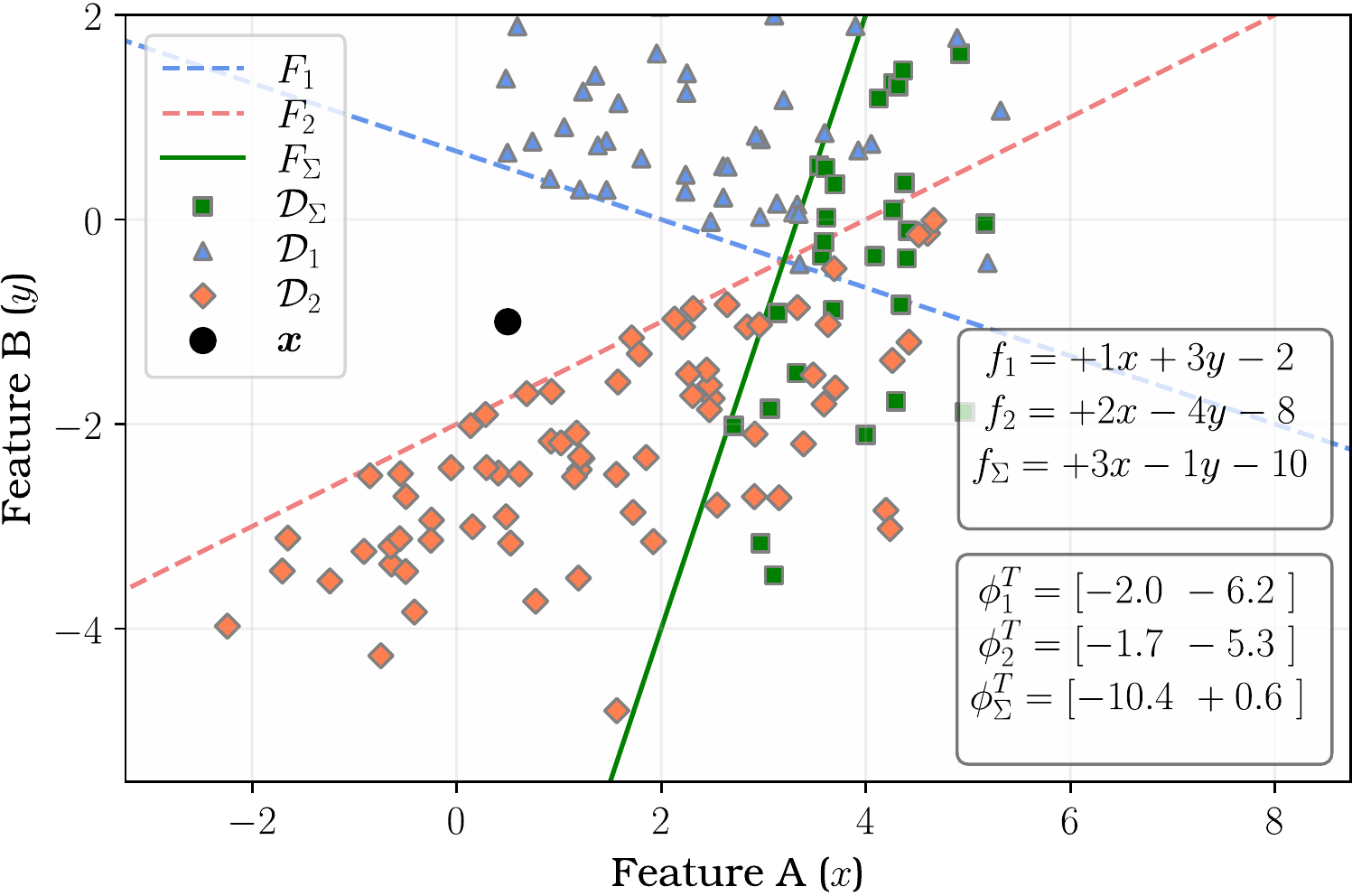}
\caption{Linearity for Counterfactual SHAP explanations: solid lines indicate the models' ($f_1$, $f_2$ and $f_{\Sigma} = f_1 + f_2$) decision boundaries ($F_1$, $F_2$ and $F_{\Sigma}$); the black point indicates the input $\x$; the coloured squares are samples of the background distribution $\D_{D-PRED}(f)$.
}
\label{fig:example_additivity}
\end{figure*}

This should not be seen as a limitation of Counterfactual SHAP, but rather as a necessary consequence of tailoring the contrastive explanations in a counterfactual way. Indeed, even using the SHAP algorithm with the background dataset given by $\D_{\text{D-PRED}}(f)$, the samples predicted differently by $f$, results in the failure of the linearity property in this sense. This can be seen in Figure~\ref{fig:example_additivity}; this figure shows two models in red and blue and the model that results from their sum in green. For each model, we show points from the data that are predicted differently by each of these models. All of these points are counterfactual for their respective model, but the points which are counterfactual for the blue model (say) are not necessarily counterfactual for the red or green models.


\subsection{Computational complexity}

As one can deduce from Algorithm~\ref{algo:cshap}, the computation time of CF-SHAP is simply given by the the sum of the time to compute the counterfactual explanations, the Shapley values and the trends.
$$
T(n) = T_{C}(n) + T_{SHAP}(n) + T_{Trends}(n) = O(n)
$$
where $n$ is the number of (counterfactual) points in the background dataset.

The computation time of the interventional variant of TreeSHAP (that we use in this paper) depends linearly on the number of samples in the background distribution \cite{Lundberg2020Trees}, i.e., $T_{SHAP}(n) = O(n)$.
The computation time of the counterfactuals explanations depends on the counterfactual generation technique that is being used. In the case of $k$-nearest neighbours, the overall computation time depends linearly on the number of neighbours generated, i.e., once again $T_{K-NN}(n) = O(n)$.
The computation of the trends is also linear with respect to the number of counterfactual explanations since it is a simple average, therefore $T_{Trends}(n) = O(n)$.

\subsection{Execution Time: Experiments}\label{appendix:exectime_exp}
To experimentally test the execution time of the explanation techniques we recorded the (average) time taken by each explanation technique to generate a single explanation. 

\textbf{Experiment setup}.
In particular, in order to estimate the execution time for a single explanation, we run each explanation algorithm on all the samples in the dataset for a time $t > 0.1$ seconds; we then divide $t$ by the number of explanations that the method computed in such time frame. To account for error introduced by the OS scheduler we run the experiments $10$ times and, since all the explanations techniques that we experimented with are deterministic, we take the minimum execution time.

\begin{table}
    \begin{small}
    \begin{center}
\begin{tabular}{rlll}
\hline
 &                HELOC &                  LC &                 WQ \\
\hline
\multicolumn{3}{c}{\textsf{\textsc{Counterfactual SHAP}}}\\
\hline
CF-SHAP $100$-NN  &  $646 \mu s$ &  $2,646 \mu s$ &  $620 \mu s$ \\
CF-SHAP $10$-NN  &  $189 \mu s$ &    $419 \mu s$ &  $165 \mu s$ \\
\hline
\multicolumn{3}{c}{\textsf{\textsc{Baselines}}} \\
\hline
SHAP D-LAB (n = 100)  &  $619 \mu s$ &  $3,024 \mu s$ &  $652 \mu s$ \\
SHAP D-PRED (n = 100) &  $597 \mu s$ &  $3,015 \mu s$ &  $640 \mu s$ \\
SHAP TRAIN (n = 100)  &  $380 \mu s$ &  $2,209 \mu s$ &  $408 \mu s$ \\
\end{tabular}

    \end{center}
    \protect\caption{
        \EB{Execution time of different explanations techniques. We report the (average) execution time to generate the explanation of a single sample. $n$ represents the number of points in the random sample drawn from the background distribution. Refer to Appendix~\ref{appendix:exectime_exp} for more details about the setup.}
    }
    \label{table:performance}
    \end{small}
\end{table}

\textbf{Results}. 
The results are reported in Table~\ref{table:performance}. The main findings are reported as follows.
\begin{itemize}\setlength\itemsep{0em}
    \item The impact of $K$-nearest neighbour computation on the Counterfactual SHAP values computation is minimal. For example, we can see from Table~\ref{table:performance} that there is only a $4.3\%$ increase in execution time of CF-SHAP 100-NN with respect to that of SHAP D-PRED ($n=100$).
    \item The execution time of (Tree-)SHAP scales linearly with the size of the background dataset confirming the theoretical results in \cite{Lundberg17}. This means that explanations techniques using the full training or variants thereof (SHAP D-LAB/PRED) have a considerably larger run-times than other techniques. We remark that, in practise, to reduce the execution time bottle-neck, such distribution are replaced with others approximating them, e.g., a random sample or of k-means medoids.
\end{itemize}

\begin{table*}[!t]
    \begin{small}
    \begin{center}
\begin{tabular}{rr|lllll|lllll|lllll}
\toprule
            &       & \multicolumn{15}{c}{\textsc{Counterfactual-Ability Improvement (\%)}}\\
  $K$ & & \multicolumn{5}{c}{HELOC} & \multicolumn{5}{c}{Lending Club} & \multicolumn{5}{c}{Wine Quality} \\
      & $k=$ &                  $1$ &                 $2$ &                 $3$ &                 $4$ &                 $5$ &                  $1$ &                 $2$ &                 $3$ &                  $4$ &                  $5$ &                  $1$ &                  $2$ &                  $3$ &                  $4$ &                  $5$ \\
\midrule
    1 &        &       \!\!$-8.6$\!\! &      \!\!$-6.4$\!\! &     \!\!$-18.1$\!\! &     \!\!$-26.7$\!\! &     \!\!$-31.7$\!\! &      \!\!$-28.0$\!\! &     \!\!$-13.8$\!\! &      \!\!$-5.0$\!\! &        \!\!$2.6$\!\! &        \!\!$8.0$\!\! &       \!\!$-0.4$\!\! &       \!\!$11.5$\!\! &        \!\!$8.2$\!\! &        \!\!$8.2$\!\! &        \!\!$8.5$\!\! \\
    3 &        &       \!\!$-4.9$\!\! &       \!\!$2.4$\!\! &       \!\!$1.0$\!\! &      \!\!$-2.4$\!\! &      \!\!$-3.7$\!\! &      \!\!$-19.2$\!\! &      \!\!$-4.2$\!\! &       \!\!$3.8$\!\! &        \!\!$8.8$\!\! &       \!\!$18.3$\!\! &        \!\!$6.1$\!\! &       \!\!$14.3$\!\! &       \!\!$15.0$\!\! &       \!\!$15.2$\!\! &       \!\!$16.1$\!\! \\
    5 &        &       \!\!$-3.1$\!\! &       \!\!$3.9$\!\! &       \!\!$4.2$\!\! &       \!\!$2.6$\!\! &       \!\!$2.4$\!\! &      \!\!$-16.6$\!\! &      \!\!$-0.6$\!\! &       \!\!$6.6$\!\! &       \!\!$10.2$\!\! &       \!\!$20.3$\!\! &        \!\!$8.2$\!\! &       \!\!$17.2$\!\! &       \!\!$17.0$\!\! &       \!\!$16.7$\!\! &       \!\!$18.7$\!\! \\
   10 &        &       \!\!$-1.9$\!\! &  \!\!$\bm{6.4}$\!\! &       \!\!$7.0$\!\! &       \!\!$4.7$\!\! &       \!\!$5.3$\!\! &      \!\!$-14.0$\!\! &  \!\!$\bm{2.4}$\!\! &  \!\!$\bm{6.7}$\!\! &  \!\!$\bm{13.0}$\!\! &  \!\!$\bm{21.6}$\!\! &        \!\!$9.6$\!\! &  \!\!$\bm{19.0}$\!\! &       \!\!$17.8$\!\! &  \!\!$\bm{18.9}$\!\! &  \!\!$\bm{19.1}$\!\! \\
   20 &        &       \!\!$-1.0$\!\! &  \!\!$\bm{6.4}$\!\! &  \!\!$\bm{8.0}$\!\! &  \!\!$\bm{5.4}$\!\! &       \!\!$5.4$\!\! &      \!\!$-12.2$\!\! &       \!\!$2.1$\!\! &       \!\!$6.2$\!\! &       \!\!$12.4$\!\! &       \!\!$19.5$\!\! &  \!\!$\bm{10.5}$\!\! &       \!\!$18.0$\!\! &  \!\!$\bm{18.3}$\!\! &       \!\!$18.8$\!\! &       \!\!$18.5$\!\! \\
   50 &        &       \!\!$-1.0$\!\! &       \!\!$5.2$\!\! &       \!\!$7.2$\!\! &       \!\!$5.3$\!\! &       \!\!$6.2$\!\! &       \!\!$-9.5$\!\! &  \!\!$\bm{2.4}$\!\! &       \!\!$5.7$\!\! &       \!\!$11.2$\!\! &       \!\!$18.4$\!\! &        \!\!$9.4$\!\! &       \!\!$18.3$\!\! &       \!\!$17.8$\!\! &       \!\!$15.3$\!\! &       \!\!$13.7$\!\! \\
  100 &        &       \!\!$-1.0$\!\! &       \!\!$3.7$\!\! &       \!\!$6.3$\!\! &       \!\!$5.1$\!\! &  \!\!$\bm{6.4}$\!\! &       \!\!$-7.8$\!\! &       \!\!$1.6$\!\! &       \!\!$5.0$\!\! &        \!\!$9.8$\!\! &       \!\!$17.1$\!\! &        \!\!$7.9$\!\! &       \!\!$15.3$\!\! &       \!\!$15.2$\!\! &       \!\!$12.0$\!\! &        \!\!$9.7$\!\! \\
  250 &        &       \!\!$-1.1$\!\! &       \!\!$2.0$\!\! &       \!\!$4.7$\!\! &       \!\!$4.6$\!\! &       \!\!$6.0$\!\! &       \!\!$-4.4$\!\! &       \!\!$2.1$\!\! &       \!\!$3.9$\!\! &        \!\!$8.9$\!\! &       \!\!$13.2$\!\! &        \!\!$4.6$\!\! &        \!\!$8.6$\!\! &        \!\!$5.2$\!\! &        \!\!$0.9$\!\! &       \!\!$-0.1$\!\! \\
  500 &        &       \!\!$-1.2$\!\! &       \!\!$1.4$\!\! &       \!\!$3.7$\!\! &       \!\!$4.0$\!\! &       \!\!$5.8$\!\! &       \!\!$-2.7$\!\! &       \!\!$1.2$\!\! &       \!\!$3.2$\!\! &        \!\!$8.1$\!\! &       \!\!$10.6$\!\! &        \!\!$1.2$\!\! &       \!\!$-1.6$\!\! &       \!\!$-8.2$\!\! &      \!\!$-14.8$\!\! &      \!\!$-18.9$\!\! \\
 1000 &        &  \!\!$\bm{-0.7}$\!\! &       \!\!$0.5$\!\! &       \!\!$1.6$\!\! &       \!\!$2.2$\!\! &       \!\!$3.7$\!\! &  \!\!$\bm{-1.2}$\!\! &       \!\!$0.8$\!\! &       \!\!$1.8$\!\! &        \!\!$5.5$\!\! &        \!\!$7.7$\!\! &        \!\!$0.7$\!\! &      \!\!$-13.2$\!\! &      \!\!$-26.6$\!\! &      \!\!$-34.1$\!\! &      \!\!$-37.0$\!\! \\
\bottomrule
\end{tabular}
    \end{center}
    \protect\caption{
        \EB{Counterfactual-ability improvement (as defined in Section~\ref{sec:experiments}) of CF-SHAP with respect to SHAP D-PRED (higher is better) for different datasets when varying the number of nearest neighbors ($K$) selected as counterfactuals and the number of features allowed to change ($k$).}
    }
    \label{table:increasing_k}
    \end{small}
\end{table*}

\section{Changing the number of nearest neighbours ($K$)}
\label{appendix:increasingk}

\ED{
The techniques based on $K$-nearest neighbours that we used to compute counterfactuals in our experiments have as main parameter the number of neighbours $K$. In this appendix we will provide the results showing how the counterfactual-ability of explanations changes when varying $K$.
We remark, as already noted in Section~\ref{sec:experiments}, that to allow for a fair comparison with Counterfactual SHAP, we set $K=100$ when running our main experiments. In fact, by default, \texttt{shap} randomly sample $100$ points from the a background dataset it has provided with.

\textbf{Experimental Setup}. We measured the improvement in {counterfactual-ability} as described in Section~\ref{sec:experiments} over $4,000$ explanations. For this experiment we chose to compare against the toughest baseline to beat: SHAP D-PRED. We varied the number of neighbors used by CF-SHAP from $1$ to $1000$ and recorded the results.

\textbf{Results}. Table~\ref{table:increasing_k} reports the improvement in counterfactual-ability of CF-SHAP when compared to SHAP D-PRED.

In particular, we note that:
\begin{itemize}\setlength\itemsep{0em}
    \item Depending on the dataset the best choice for the hyper-parameter $K$ of $K$-NN ranges from $K=10$ to $K=100$.
    \item Increasing the number of neighbours over $100$ does not result in an improvement in counterfactual-ability. This matches with the intuition that increasing the number of counterfactuals included in the background dataset means that the explanation will be ``less tailored'' to the specific user (associated with the input) $\x$.
    \item Decreasing the number of neighbours under $10$ does not result in an improvement in counterfactual-ability. This again matches with the intuition that proving (too) few points as background distribution makes the explanation less robust because it reduces the ability of CF-SHAP to describe the decision boundary. In particular, we can see how providing a single counterfactual point results in a very sharp fall in counterfactual-ability of the explanation.
\end{itemize}
}

\section{Robustness to Different Settings}\label{appendix:moreexperiments}
\ED{

To show the robustness of our evaluation we report additional results for the experiments on counterfactual-ability improvement and plausibility improvement under different action functions and cost functions. In the main text we

\EF{
\textbf{Experimental Setup}. We run the same experiments for the counterfactual-ability and plausibility as reported in Section~\ref{sec:experiments} using alternative definitions of action function and cost function that we have introduced in the paper. In particular in Section \ref{sec:experiments} we reported the results for the improvement in counterfactual-ability (Figure~\ref{fig:exp_costsL1}) and the improvement in plausibility (Figure~\ref{fig:exp_plausaL1}) under the assumption of:
\begin{itemize}\setlength\itemsep{0em}
\item random recourse (i.e. using action function $\arandk$) and total quantile shift cost (i.e., using cost function $\cf_{L1}$).
\end{itemize}
In this appendix we report the results under the following alternative assumptions:
\begin{itemize}\setlength\itemsep{0em}
\item random recourse (action function $\arandk$) and quantile shift cost with L2 norm (i.e., using cost function $\cf_{L2}$);
\item proportional recourse (i.e., using action function $\apropk$) and total quantile shift cost (i.e., using cost function $\cf_{L1}$);
\item proportional recourse (i.e., using action function $\apropk$) and total quantile shift cost under L2 norm (i.e., using cost function $\cf_{L2}$).
\end{itemize}

\textbf{Results}. Figure~\ref{fig:costs_alternative} and \ref{fig:plausibi_alternative} shows the results for the improvement in counterfactual-ability and plausibility, respectively, under these different assumptions. We note that both the counterfactual-ability and the plausibility experiments results are highly robust to the choice of action and cost function. This suggests that the CF-SHAP performs better than the baselines also when changing the underlying assumptions on how users may act on the explanation (action function) and how users measure the cost of the recourse (cost function).}

\begin{figure*}[t]
\centering
\includegraphics[width=.85\textwidth]{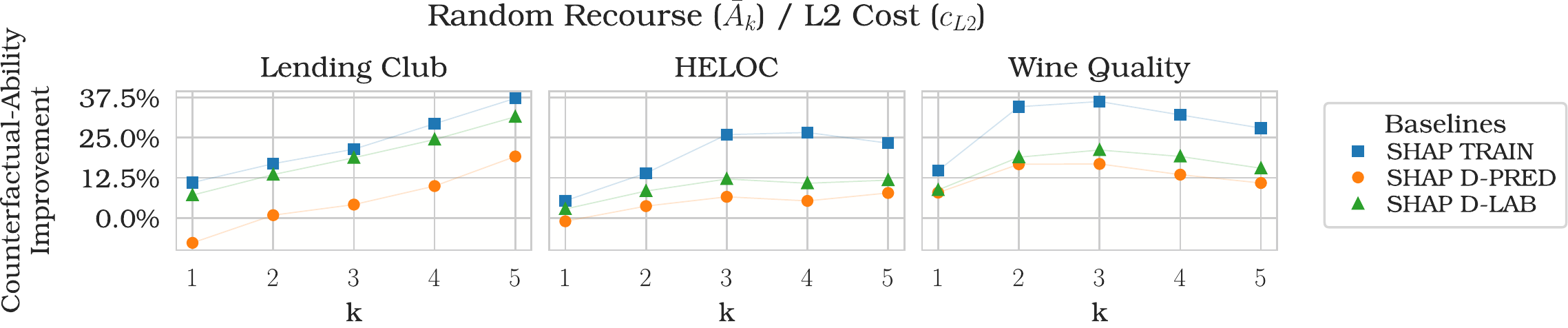}
\includegraphics[width=.85\textwidth]{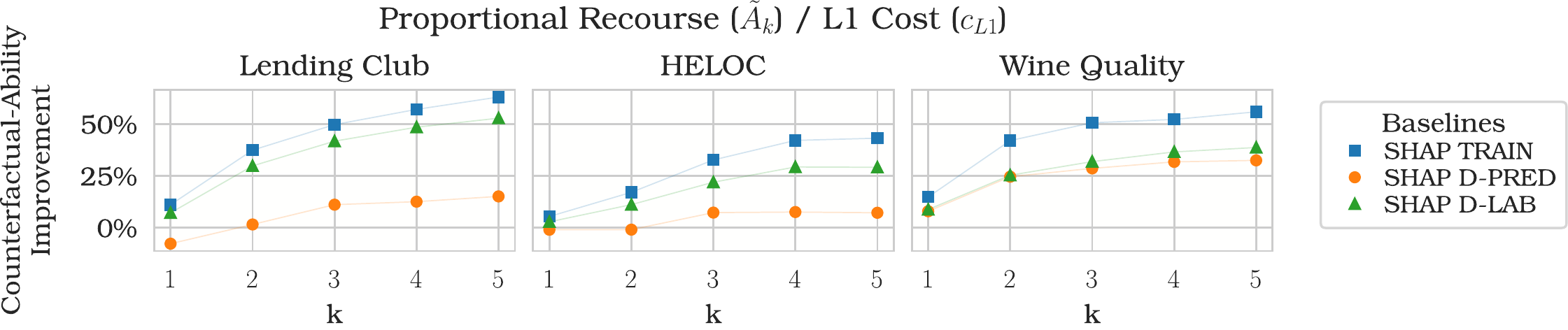}
\includegraphics[width=.85\textwidth]{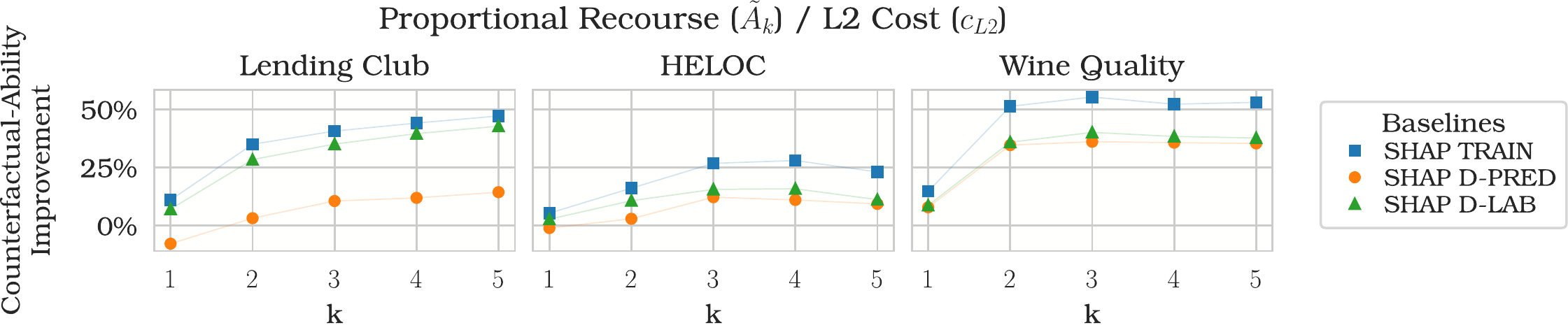}
\caption{Improvement in counterfactual-ability under different assumptions (i.e., using alternative definitions of action function and cost function). This is the equivalent of Figure~\ref{fig:exp_costsL1} under different assumptions. See Appendix~\ref{appendix:moreexperiments} for more details.}
\label{fig:costs_alternative}
\end{figure*}
\begin{figure*}
\centering
\includegraphics[width=.85\textwidth]{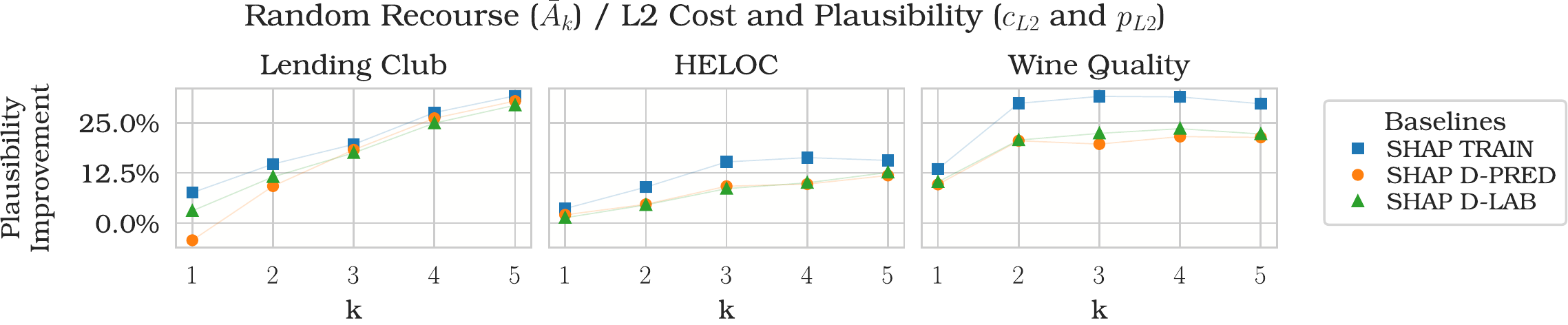}
\includegraphics[width=.85\textwidth]{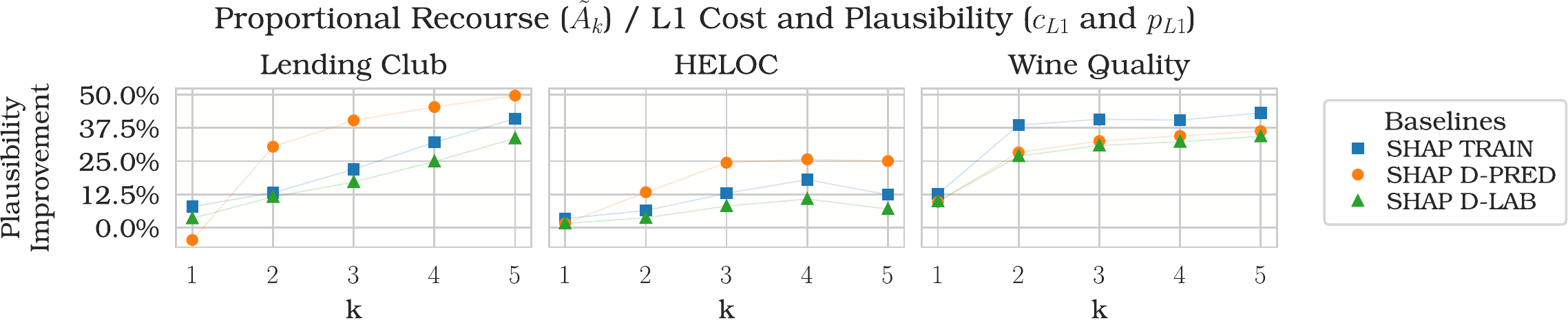}
\includegraphics[width=.85\textwidth]{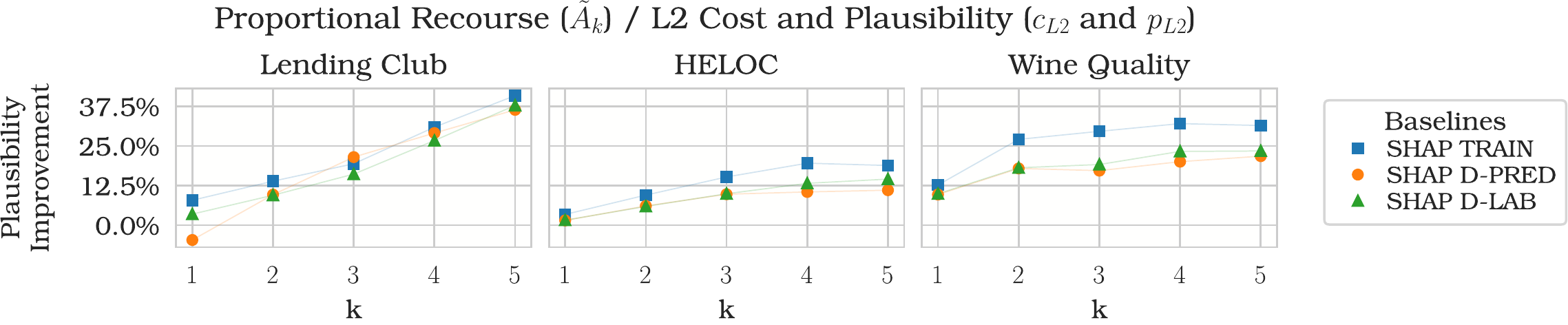}
\caption{Improvement in plausibility under different assumptions (i.e., using alternative definitions of action function and cost function). This is the equivalent of Figure~\ref{fig:exp_plausaL1} under different assumptions. See Appendix~\ref{appendix:moreexperiments} for more details.}
\label{fig:plausibi_alternative}
\end{figure*}
}

\end{document}